\documentclass[runningheads]{llncs}

\usepackage{accv}

\usepackage{accvabbrv}

\usepackage{enumitem}
\usepackage[table]{xcolor}
\usepackage{multirow}
\usepackage{threeparttable}
\usepackage{algorithm,algorithmicx,algpseudocode}
\usepackage{makecell}
\usepackage{amsmath}
\usepackage{listings}
\usepackage{graphicx}
\usepackage{booktabs}

\newcommand{\tablestyle}[2]{\setlength{\tabcolsep}{#1}\renewcommand{\arraystretch}{#2}\centering\footnotesize}
\definecolor{gray}{HTML}{efefef}

\usepackage[accsupp]{axessibility}  

\usepackage[pagebackref,breaklinks,colorlinks,citecolor=accvblue]{hyperref}

\usepackage{orcidlink}

\begin{document}

\title{FFNet: MetaMixer-based Efficient Convolutional Mixer Design} 

\titlerunning{FFNet}

\author{Seokju Yun \inst{1} \and
Dongheon Lee\inst{2} \and
Youngmin Ro\inst{2}
}
\authorrunning{S.~Yun et al.}

\institute{
KAIST AI
\and
University of Seoul \\
\url{https://github.com/ysj9909/FFNet}
}

\maketitle

\begin{abstract}
Transformer, composed of self-attention and Feed-Forward Network (FFN), has revolutionized the landscape of network design across various vision tasks.
While self-attention is extensively explored as a key factor in performance, FFN has received little attention in vision domains.
FFN is a versatile operator seamlessly integrated into nearly all AI models.
Recent works show that FFN functions like key-value memories.
Thus, akin to the query-key-value mechanism within self-attention, FFN can be viewed as a memory network, where the input serves as query and the two projection weights operate as keys and values, respectively.
Based on these observations, we hypothesize that the importance lies in query-key-value framework itself for competitive performance.
To verify this, we propose converting self-attention into a more FFN-like efficient token mixer with only convolutions while retaining query-key-value framework, namely \textit{FFNification}.
Specifically, FFNification replaces query-key-value interactions with large kernel convolutions and adopts GELU activation function instead of softmax.
The derived token mixer, \textit{FFNified attention}, serves as key-value memories for detecting locally distributed spatial patterns, and operates in the opposite dimension to the ConvNeXt block within each corresponding sub-operation of the query-key-value framework.
Building upon the above two modules, we present a family of Fast-Forward Networks (FFNet).
Despite being composed of only simple operators, FFNet outperforms sophisticated and highly specialized methods in each domain, with notable efficiency gains.
These results validate our hypothesis, leading us to propose “MetaMixer”, a general mixer architecture that does not specify sub-operations within the query-key-value framework.
Building on the MetaMixer framework, we also introduce a hybrid strategy that harmoniously integrates attention and FFNified attention, offering a comprehensive view of mixer design.
    
  \keywords{Efficient Vision Backbone \and Key-value Memory \and Convolution}
\end{abstract}
\section{Introduction} \label{sec:intro}

Transformer~\cite{Vaswani2017transformer} models have emerged as the dominant backbone in Natural Language Processing and have successfully extended their influence to the vision domains.
Vision Transformer (ViT)~\cite{dosovitskiy2021vit} and its follow-ups~\cite{carion2020detr, liu2021swin} have showcased promising performances in various vision tasks, posing a formidable challenge to the established Convolutional Neural Networks (CNNs)~\cite{he2016resnet}.
Main component of the Transformer, global self-attention, is specialized in capturing global contexts but incurs a quadratic computational complexity with respect to the sequence length.
Therefore, deploying this module on resource-constrained mobile devices for real-time applications proves to be much more challenging than using lightweight CNNs.
To address these issues, a lot of works have focused on modifying the self-attention module to be more vision-friendly~\cite{mehta2021mobilevit,shaker2023swiftformer} or selectively applying it to low-resolution inputs~\cite{vasu2023fastvit, yun2024shvit}.
As such, effectively leveraging self-attention under strict latency constraints requires laborious adjustments.

\begin{figure}[t]
    \centering
    \includegraphics[width=0.7\linewidth]{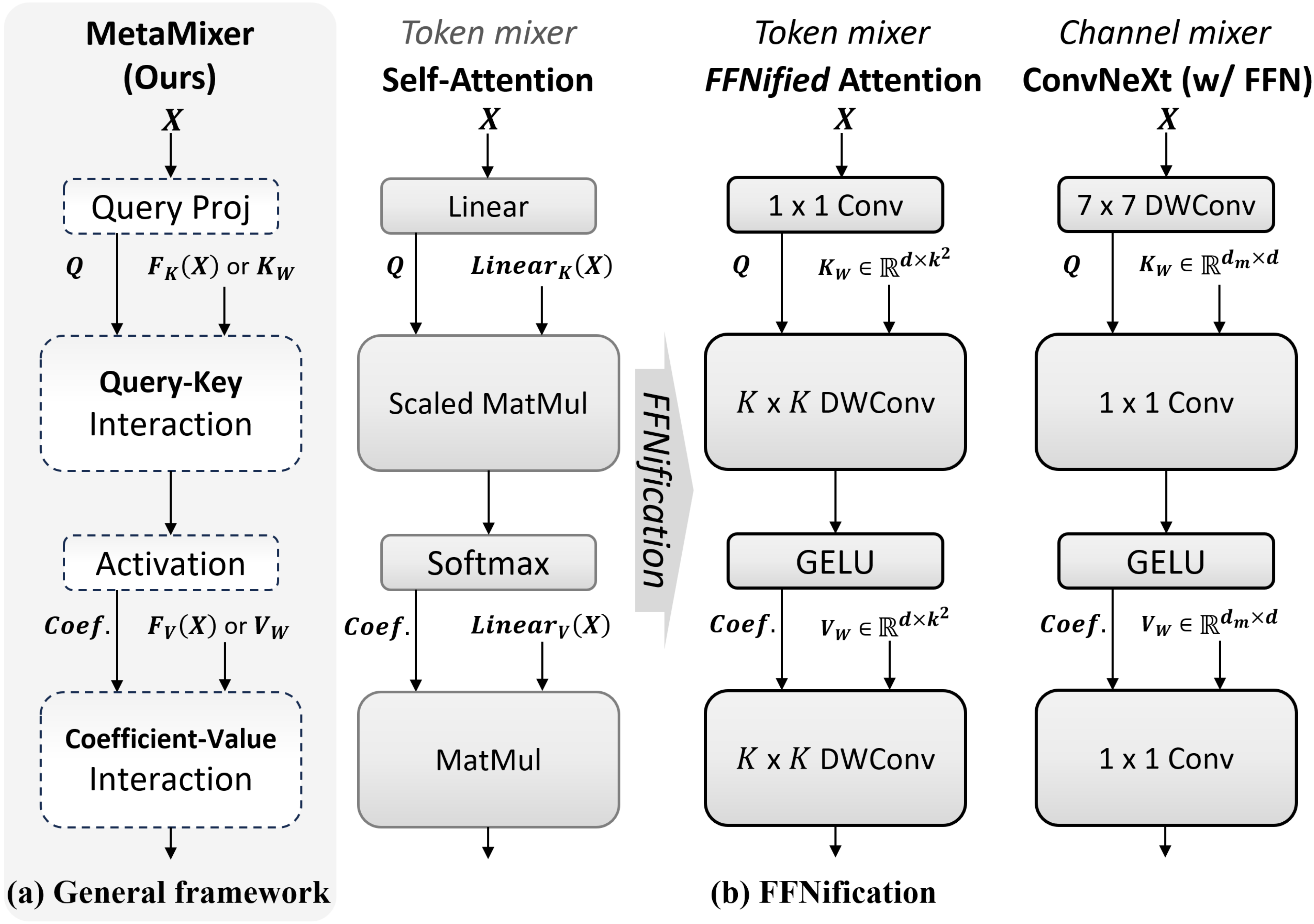}
    \vspace{-1mm}
    \caption{
    \textbf{Overview of MetaMixer.} 
  (a) \emph{MetaMixer} is derived by not specifying sub-operations within the query-key-value framework.
  We assert that the competence of Transformers primarily originates from MetaMixer, which we deem as the true \textbf{backbone} of Transformer.
  (b) To demonstrate this and propose a FFN-like efficient token mixer, we replace the inefficient sub-operations of self-attention with those from FFN within MetaMixer structure.
    }
    \label{fig:metamixer}
    \vspace{-3mm}
\end{figure}

Most models effortlessly leverage Feed-Forward Networks (FFNs), which aggregate channel information with moderate computational overhead. 
Nevertheless, in vision domains, FFN is usually excluded from analysis, although there are several motivations to consider FFNs.
For example, regardless of modality, whether generative or perception tasks, FFNs serve as a fundamental building block in transformer-based architectures, such as GPT series, ViT and its diffusion-based variants~\cite{peebles2023dit}; this ubiquity suggests that FFNs play an indispensable role in achieving competitive performance.
Furthermore, \cite{geva2020keyvalue} suggest that FFNs can be viewed as emulated neural key-value memories~\cite{sukhbaatar2019augmentingmemory, sukhbaatar2015kvmemory}, where the rows in the first projection weights act as keys correlating with input patterns, and the rows in the second projection weights serve as values that induce a distribution shift over the residual path.
Based on this intuition, if a depthwise convolution is used as query projection, ConvNeXt block~\cite{liu2022convnet} can also be interpreted as operating via a query-key-value mechanism, similar to the self-attention (see Fig.~\ref{fig:metamixer}).
We thus hypothesize that \textit{the query-key-value framework, fundamental to the most successful mixers like self-attention and ConvNeXt block, is essential as the starting point for mixer design.}

We thus abstract the query-key-value mechanism into a generalized framework, \textit{MetaMixer}, where sub-operations are not predefined, as illustrated in Fig.~\ref{fig:metamixer} (a).
We further propose to replace the inefficient sub-operations of self-attention with those of FFN within the MetaMixer framework, aiming to bridge the efficiency gap between self-attention and FFN.
We term it as \textit{FFNification}.
It consists of three aspects.
\textbf{(1) Key-Value Generation}: Self-attention generates keys and values through linear projections (dynamic computation), whereas FFN employs static weights as keys and values.
We thus adopt a static weight scheme for simple and efficient implementation.
\textbf{(2) Activation Function (Coefficient Generation)}: Softmax is costly because it involves calculating an exponent and summing across the token length, which poses challenges for parallelization~\cite{dao2022flashattention}.
Consequently, we replace the softmax with an element-wise activation function GELU~\cite{hendrycks2016gelu}.
\textbf{(3) Query-Key-Value Interactions}: The query-key and coefficient-value interactions in self-attention result in quadratic complexity $O(n^2)$ with the token length $n$, as they compute relations between a single token and all other tokens.
Instead, we explore whether these expensive calculations can be efficiently substituted by simple local aggregation like convolution.
Inspired by large kernel CNNs~\cite{liu2022convnet}, we primarily use kernel sizes of 7 or even larger.
The resulting \emph{FFNified attention} and ConvNeXt block are symmetrical in terms of mixing dimension (see Fig.~\ref{fig:metamixer} (b)), and are thus adopted as the token mixer and channel mixer, respectively.

In this paper, we validate the MetaMixer framework by conducting extensive experiments across various tasks leveraging only convolutions.
Our MetaMixer-based convolutional networks establish a strong and generally efficient baseline across a wide range of tasks and devices.
Importantly, our convolutional mixer design surpasses highly specialized task-specific mixers with minimal modifications.
Based on these results, we argue that MetaMixer stands as the pivotal mixer architecture that ensures competitive performance.
We further discuss how to integrate existing mixer and FFNified attention within the MetaMixer, taking into account the characteristics of the task and the model's data processing mechanism.
We make the following contributions:
\begin{itemize}[leftmargin=3mm, itemsep=0.5mm, topsep=1mm, partopsep=0mm]
    \item We provide an in-depth analysis of the query-key-value mechanism of FFN and abstract this mechanism into a general mixer architecture \emph{MetaMixer} that encompasses the main modules of both Transformer and ConvNeXt.
    \item  Although self-attention and FFN-based convolutional mixers can both be integrated within the MetaMixer, a significant efficiency gap remains between them. To address this, we propose a mixer design that adapts self-attention by employing FFN's sub-operations and mimicking its functionality through the use of large kernel convolutions. The resulting token mixer, termed \emph{FFNified attention}, outperforms attention-based mixers in terms of the speed-accuracy trade-off across diverse devices.
    \item Through a comprehensive comparison of our models with SOTA baselines across various modalities and tasks, we underscore the critical importance of the query-key-value framework (\emph{i.e.,} MetaMixer) and offer a strong and efficient baseline adaptable to various settings.
\end{itemize}
\section{FFNs Are Key-Value  Memories} \label{sec: FFNs_are_kvm}
Transformer block is primarily composed of self-attention module and Feed-Forward Network (FFN).
Let $X \in \mathbb{R}^{n \times d}$ denote the input matrix, we can express self-attention and FFN as follows (bias terms, multi-head mechanism, and scaling factor in self-attention are omitted for simplicity):
\begin{gather}
    Q = XW_Q, K = XW_K, V = XW_V, \\
    \mathrm{Attention}(X) = \mathrm{Softmax}(QK^{\top})V, \\
    \hspace{0.95cm} \mathrm{FFN}(X) = \mathrm{GELU}(XW_1^{\top})W_2, 
\end{gather}
where $W_1, W_2 \in \mathbb{R}^{d_m \times d}$ and $ W_Q, W_K, W_V \in \mathbb{R}^{d \times d_h}$.
As shown in the above equations, the two projection weights of FFN operate similarly to the key-value in self-attention; the slight differences are that FFN is applied along the channel axis and uses GELU~\cite{hendrycks2016gelu} instead of softmax.
Specifically, the keys in $W_1$ detect patterns in the training samples, yielding the unnormalized memory coefficients $\textbf{c} = \mathrm{GELU}(XW_1^{\top})$, referred to as the \emph{coefficient}, which are then used to integrate the values in $W_2$.

\begin{figure}[t]
    \centering
    \includegraphics[width=0.9\linewidth]{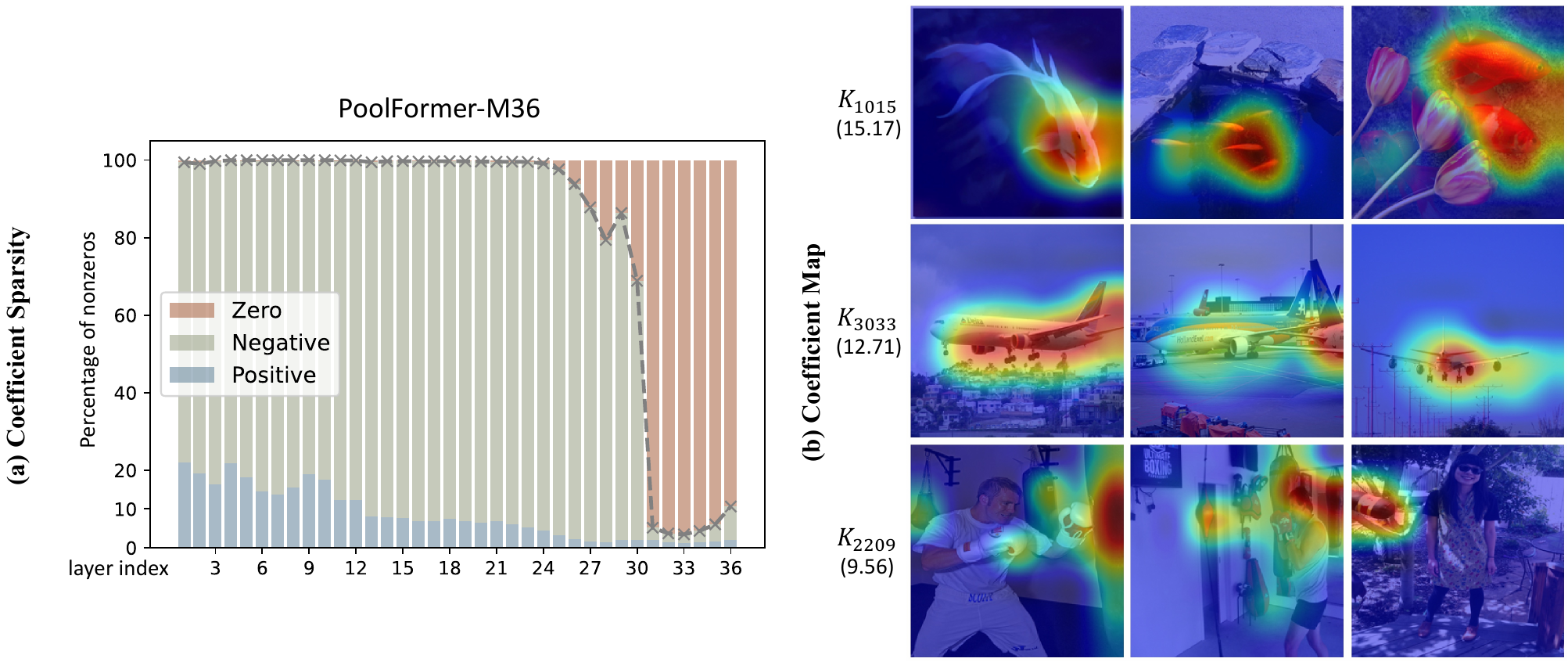}
    \vspace{-0.5mm}
    \caption{\textbf{Key-Value Mechanism of FFNs.} (a) Coefficient Sparsity: Astonishingly, the final stage shows significantly higher sparsity. To categorize numerous classes,
a small subset ($<$ 10\%) of neurons are activated. 
(b) Coefficient Map corresponding to the most activated key in the last layer: Keys specialized for each class selectively correlate at the target regions, suggesting their potential role in capturing distinctive visual features.
The numbers in parentheses indicate the average values of the coefficients.}
    \label{fig:FFNs_are_kvm}
    \vspace{-3mm}
\end{figure}

NLP literatures~\cite{meng2022locatinggpt, dai-etal-2022-knowledgen-neurons, wang2022findingskills} also delve into the knowledge encoded within FFNs through viewing them as key-value memories~\cite{sukhbaatar2015kvmemory}.
For instance, \cite{geva2020keyvalue} find that each key correlates with a range of human-interpretable patterns; \cite{dai-etal-2022-knowledgen-neurons} and \cite{meng2022locatinggpt} update specific factual associations by identifying and editing coefficients that express certain factual knowledge.
Also, from a different perspective, recent studies find that coefficient sparsity in FFNs is a prevalent phenomenon~\cite{li2022lazy, zhang2021moefication}.

While analyses of FFNs have reached maturity in NLP, the computer vision community lacks similar investigations.
We thus analyze the FFNs within PoolFormer~\cite{yu2021metaformer} from a key-value memory perspective.
The simplicity of pooling as the token mixer makes FFN the dominant computational component in terms of parameter count and capacity, highlighting its crucial role in the model's overall performance.
As shown in Fig.~\ref{fig:FFNs_are_kvm} (a), consistent with the observations in \cite{mirzadeh2024relustrikeback, li2022lazy}, layers in the final stage have $<$ 10\% activated coefficients.
Considering that each value can be interpreted as a distribution over the output prediction for each class~\cite{geva2020keyvalue}, the result is intuitive.
Also, the coefficient map examples in Fig.~\ref{fig:FFNs_are_kvm} (b) demonstrate that class-specific keys consistently correlate with target objects.
Further examples and more detailed discussions are provided in the Appendix.
\textit{These results demonstrate that FFNs are indeed key-value memories, aggregating input patterns through interpretable inner workings.
So, it is noteworthy that the two main modules of Transformer operate within the query-key-value framework.}
\section{Proposed Method}

\subsection{MetaMixer Framework} \label{sec:metamixer}
For the first time, we introduce the core concept “MetaMixer”.
As shown in Fig.~\ref{fig:metamixer}, MetaMixer is a flexible structure that does not specify sub-operations in the \textbf{query-key-value framework}~\cite{sukhbaatar2015kvmemory}.
First, the query $q$ is generated by applying a projection layer to the input. 
Additionally, key-value pairs are either computed similarly to the query or randomly initialized as memory vectors, analogous to FFNs.
We compute the match between $q$ and each key $k_i$ using a compatibility function (e.g., dot-product or convolution), followed by an activation function:
\begin{gather}
    c_i = \mathrm{Activation}(f(q ,k_i)), \label{equ:coeffcient}
\end{gather}
where $f(\cdot)$ is a compatibility function that computes the similarity between the query and the corresponding key (referred to as \emph{query-key interaction}); $\mathrm{Activation}(\cdot)$  
generates the coefficient $c$ on the values $v$.
The output is then computed by an aggregation function $g(c ,v)$ that sums the values, each weighted by the coefficient from Eq.~\ref{equ:coeffcient}.
We refer to the process of aggregating values with their associated coefficients as \emph{coefficient-value interaction}.

\begin{figure*}[t]
    \centering
    \includegraphics[width=0.8\linewidth]{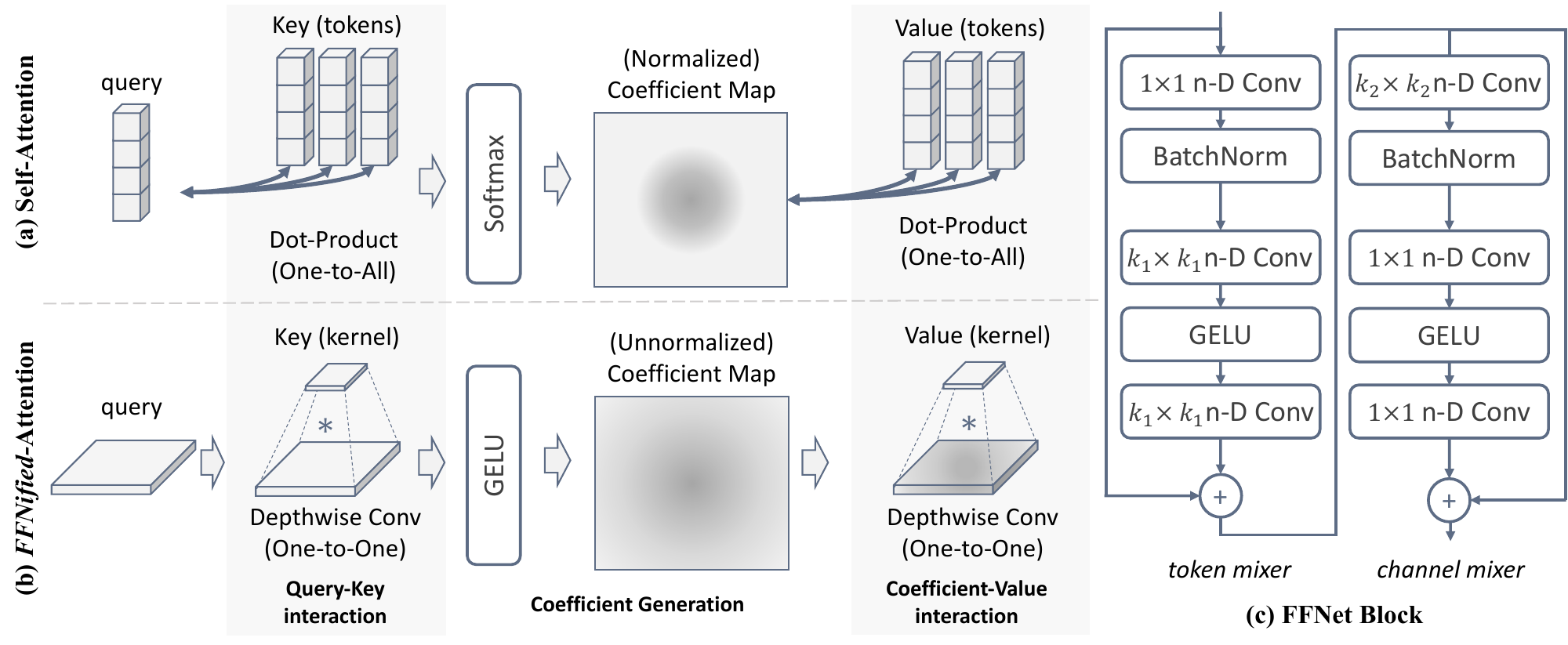}
    \vspace{-0.5mm}
    \caption{\textbf{Overview of FFNification and Fast-Forward Network block.} (a-b) Comparison between self-attention and FFNified attention; (c) Our mixer design easily adapts by selecting the convolution type and kernel size based on the modality.}
    \label{fig:FFNification}
    \vspace{-3mm}
\end{figure*}

MetaMixer, as a general mixer architecture, can be instantiated as various well-established mixers by specifying suboperations (query-key-value generation, query-key-value interactions, and activation).
This nature confirms the capabilities of MetaMixer's modular design.
 We provide a detailed illustration of MetaMixer's generality in the Appendix.
\subsection{FFNification} \label{sec:ffnification}

Although we have shown that self-attention and FFN can be interpreted within the query-key-value framework, we note that self-attention  requires labor-intensive tuning and is less efficient in vision domains compared to FFN-based convolutional mixers~\cite{liu2022convnet, tan2019efficientnet}.
Hence, to incorporate the robust computational benefits of FFNs and further validate the significance of the MetaMixer, we perform architectural surgeries that replace computationally intensive operations in self-attention with efficient alternatives.
We refer to this process as \emph{FFNification}.
Importantly, we show that replacing the expensive pairwise dot-product in query-key/coefficient-value interactions with depthwise convolution is feasible, and that softmax is no longer an indispensable component.

The most critical aspect of the FFNification is employing depthwise convolution for query-key-value interactions.
In self-attention, as inter-token module, each query (token) interacts with all instance-specific key-value pairs through dot-product operations.
In contrast, with depthwise convolution, a query (channel) interacts with static key-value (kernel) in a sliding window manner.
\emph{FFNified attention}, therefore, utilizes key-value pairs to compute local similarities and aggregate information (see Fig.~\ref{fig:FFNification} (a-b)). 

Leveraging convolutional interaction is not only efficient but also benefits from inductive biases such as locality and weight sharing across positions, which reduces model complexity and aids in learning translationally equivalent representations.
To enlarge the receptive field, we mainly utilize a 7$\times$7 kernel size for image recognition tasks, while for other tasks, the kernel size is adjusted based on the characteristics of the modality and task.
Moreover, to enhance performance without inference overhead, we leverage a structural re-parameterization~\cite{ding2021repvgg}, training an additional small kernel branch alongside a large kernel, which are subsequently merged into the large kernel during inference.

While softmax self-attention has long been the \emph{de facto} choice, its suitability in local computation like convolution warrants reconsideration.
The exponential normalization in softmax tends to be highly centralized in a small number of values~\cite{ ma2022fcvit, shen2023reluformer}, which means convolution with a local receptive field may fail to achieve sufficient context aggregation.
In contrast, GELU do not suffer from this issue. 
Therefore, we adopt GELU in our settings.

The derived \emph{FFNified attention} consists solely of hardware-friendly operations. Furthermore, within the MetaMixer, the FFNified attention and ConvNeXt block are symmetrically positioned to operate in opposite dimensions across each sub-operation (see Fig.~\ref{fig:metamixer} (b)). Consequently, we employ FFNified attention and the ConvNeXt block as token mixer and channel mixer, respectively.
This arrangement facilitates efficient and compact feature extraction.
\subsection{FFNet as a General Backbone}

We introduce Fast-Forward Network (FFNet), a new network family that achieves competitive results compared to sophisticated methods on diverse benchmarks, utilizing extremely simple operators based on MetaMixer.
The basic block of our FFNet is illustrated in Fig.~\ref{fig:FFNification} (c).
Moreover, our model consists solely of hardware-friendly operations, ensuring high efficiency across various devices.

Our mixer design can be easily adapted to various modalities with minimal modifications. For image data, we utilize 2D depthwise convolution, while for time series analysis, 1D depthwise convolution is employed. For point clouds, we leverage 3D submanifold convolution~\cite{graham2018submanifold}, which introduces the hash table to retrieve input indices and processes only the non-empty elements, thus preserving the geometrical structure and maintaining low computation costs.
For more in-depth details regarding the specific adaptations for each task, please refer to the corresponding subsections.
\section{Experiments} \label{sec:exp}

To validate the inherent effectiveness of the MetaMixer framework and its resulting convolutional mixer design, we conduct a comprehensive comparison of our Fast-Forward Network (FFNet) with state-of-the-art baselines across a range of tasks, such as image classification, object detection, 2D instance/semantic segmentation, super-resolution, 3D semantic segmentation, and long-term time series forecasting (the 2D semantic segmentation experiments are presented in the Appendix).
Lastly, building upon MetaMixer’s holistic view of mixer design, we harmoniously integrate self-attention and FFNified attention to improve the performance of ImageNet generation models.

\begin{table*}[!t]
    \begin{minipage}{\textwidth}
    \centering
    \caption{
        \textbf{ImageNet-1k classification.} 
        “-” denotes that model could not be reliably exported by either TensorRT or Core ML Tools.
        }\label{tab:imagenet}
        \tablestyle{4.5pt}{0.7}
        \scriptsize
        \begin{tabular}{lcccccc}\toprule
  \multirow{3}[3]{*}{Model} & Image & Params & FLOPs & GPU & Mobile & Top-1 \\ 
  & Size &  &  & Throu. & Latency & Acc. \\
   & (Px) & (M) & (G) & \scalebox{0.9}{(imgs/s)} & (ms) & (\%) \\
  \midrule
  EfficientNet-B1~\cite{tan2019efficientnet} & $240^2$ & 7.8 & 0.7 & 1115 & 1.8 & 79.1 \\
  FastViT-SA12~\cite{vasu2023fastvit} & $256^2$ & 10.9 & 1.9 & 1274 & 2.0 & 80.6 \\
  EdgeViT-S~\cite{pan2022edgevits} & $224^2$ & 11.1 & 1.9 & 621 & 5.0 &81.0 \\
  Swin-T~\cite{liu2021swin} & $224^2$ & 29.0 & 4.5 & - & 24.8 & \textbf{81.3} \\
  FasterNet-S~\cite{fasternet} &$224^2$ &31.1 &4.6 &1040 &2.5 &\textbf{81.3} \\
\cellcolor[HTML]{efefef}\textbf{FFNet-1} &\cellcolor[HTML]{efefef} $256^2$ & \cellcolor[HTML]{efefef}13.7 & \cellcolor[HTML]{efefef}2.9 & \cellcolor[HTML]{efefef}\textbf{1430} & \cellcolor[HTML]{efefef}\textbf{1.8} & \cellcolor[HTML]{efefef}\textbf{81.3} \\
  \midrule
  ConvNeXt-T~\cite{liu2022convnet} & $224^2$ &29.0 & 4.5 & \textbf{952} & 3.7 & 82.1 \\
  SMT-T~\cite{smt} &$224^2$ &11.5 & 2.4 & 802 & 23.2 & 82.2 \\
  RMT-T~\cite{fan2024rmt} &$224^2$ &14.0 &2.5 &399 &- &82.4 \\
  FastViT-SA24~\cite{vasu2023fastvit} & $256^2$ & 20.6 & 3.8 & 596 & \textbf{3.0} & 82.6 \\
  HorNet-T$_{7\times 7}$~\cite{rao2022hornet} & $224^2$ &22.0 & 4.0 & 445 & 3.4 & 82.8 \\
  EfficientNet-B4~\cite{tan2019efficientnet} & $380^2$ & 19.3 & 4.2 & 652 & 5.2 & 82.9 \\
  Swin-S~\cite{liu2021swin} & $224^2$ & 50.0 & 8.7 & - & 29.3 & \textbf{83.0} \\ 
  \cellcolor[HTML]{efefef}\textbf{FFNet-2} & \cellcolor[HTML]{efefef}$256^2$ & \cellcolor[HTML]{efefef}26.9 & \cellcolor[HTML]{efefef}6.0 & \cellcolor[HTML]{efefef}776 & \cellcolor[HTML]{efefef}3.1 & \cellcolor[HTML]{efefef}82.9 \\
   \midrule 
   MOAT-0~\cite{yang2022moat} & $224^2$ & 27.8 & 5.7 & - & 5.1 & 83.3 \\
   iFormer-S~\cite{zhu2023biformer} & $224^2$ &20.0 & 4.8 & 500 & 27.3 & 83.4 \\
  Swin-B~\cite{liu2021swin} & $224^2$ &87.8 & 15.4 & - & 33.5 & 83.5 \\
  CMT-S~\cite{guo2022cmt}& $224^2$ &25.1 & 4.0 & 407 & 11.3 & 83.5\\
  FasterNet-L~\cite{fasternet} &$224^2$ &93.5 &15.5 &498 &6.8 &83.5 \\
  EfficientNet-B5~\cite{tan2019efficientnet} & $456^2$ & 30.0 & 9.9 & 439 & 11.0 & 83.6 \\
  SMT-S~\cite{smt} & $224^2$ & 20.5 & 4.7 & 421 & 41.5 & 83.7 \\
  HorNet-S$_{7\times 7}$~\cite{rao2022hornet} & $224^2$ & 49.5 & 8.8 & 351 & 5.4 & 83.8 \\
  ConvNeXt-B~\cite{liu2022convnet} & $224^2$ &88.6 & 15.4 & 399 & 8.1 &83.8 \\
  BiFormer-S~\cite{zhu2023biformer} & $224^2$ & 26.0 & 4.5 & 162 & - & 83.8 \\
  EfficientNetV2-S~\cite{tan2021efficientnetv2} & $384^2$ & 21.5 & 8.8 & 298 & 4.5 & \textbf{83.9} \\
  FocalNet-B~\cite{yang2022focalnet} & $224^2$ &88.7 & 15.4 & 287 & 10.0 & \textbf{83.9} \\
  FastViT-MA36~\cite{vasu2023fastvit} & $256^2$ & 42.7 & 7.9 & 422 & 4.7 & \textbf{83.9} \\
  \cellcolor[HTML]{efefef}\textbf{FFNet-3} & \cellcolor[HTML]{efefef}$256^2$ & \cellcolor[HTML]{efefef}48.3 & \cellcolor[HTML]{efefef}10.1 & \cellcolor[HTML]{efefef}\textbf{540} & \cellcolor[HTML]{efefef}\textbf{4.5} & \cellcolor[HTML]{efefef}\textbf{83.9} \\
  \midrule
  RMT-S~\cite{fan2024rmt} &$224^2$ &27.0 &4.5 &239 &- &84.1 \\
  DaViT-S~\cite{ding2022davit} & $224^2$ & 49.7 & 8.8 & 177 & 16.4 & 84.2 \\
  HorNet-B$_{7\times 7}$~\cite{rao2022hornet} & $224^2$ & 87.0 & 15.6 & 280 & 8.5 & 84.2 \\
  ConvNeXt-L~\cite{liu2022convnet} & $224^2$ &198.0 & 34.4 & 218 & 16.5 & 84.3 \\
  MogaNet-B~\cite{li2024moganet}& $224^2$ & 44.0 & 9.9 & 222 & \textbf{8.2} & 84.3 \\
  CMT-B~\cite{guo2022cmt} & $256^2$ &45.7 & 9.3 & 294 & 13.2 & \textbf{84.5}\\
  Swin-B~\cite{liu2021swin} & $384^2$ & 88.0 & 47.1 & - & 181.6 & \textbf{84.5} \\
  MaxViT-S~\cite{tu2022maxvit} &$224^2$ &69.0 & 11.7 & 141 &188.1 &\textbf{84.5} \\
  \cellcolor[HTML]{efefef}\textbf{FFNet-3} & \cellcolor[HTML]{efefef}$384^2$ & \cellcolor[HTML]{efefef}48.3 & \cellcolor[HTML]{efefef}22.8 & \cellcolor[HTML]{efefef}\textbf{387} & \cellcolor[HTML]{efefef}9.1 & \cellcolor[HTML]{efefef}\textbf{84.5}\\
  \midrule
  MOAT-0~\cite{yang2022moat} & $384^2$ & 27.8 & 18.2 & - & 25.2 & 84.6 \\
  iFormer-S~\cite{zhu2023biformer} & $384^2$ &20.0 & 16.1 & \textbf{334} & 80.3 & 84.6 \\
  FasterViT-3\cite{hatamizadeh2024fastervit} &$224^2$ &159.3 &18.2 &- &60.8 &84.9 \\
  RMT-B~\cite{fan2024rmt} &$224^2$ &54.0 &9.7 &162 &- &85.0 \\
  NFNet-F2~\cite{brock2021nfnet} & $352^2$ & 193.8 & 62.6 & 194 & 27.0 & 85.1 \\
  MogaNet-XL~\cite{li2024moganet} & $224^2$ & 181.0 &34.5 & 118 & 22.3 & 85.1 \\
  ConvNeXt-B~\cite{liu2022convnet} & $384^2$ & 88.6 & 45.0 & 229 & 21.3 &85.1 \\
  MaxViT-T~\cite{tu2022maxvit}&$384^2$ &31.0 &17.7 &149 & 247.6 & 85.2 \\
  HorNet-B$_{7\times 7}$~\cite{rao2022hornet}& $384^2$ & 87.0 & 45.8 & 175 & 21.3 & \textbf{85.3} \\
  \cellcolor[HTML]{efefef}\textbf{FFNet-4} & \cellcolor[HTML]{efefef}$384^2$ & \cellcolor[HTML]{efefef}79.2 & \cellcolor[HTML]{efefef}43.1 & \cellcolor[HTML]{efefef}282 & \cellcolor[HTML]{efefef}\textbf{15.2} & \cellcolor[HTML]{efefef}\textbf{85.3} \\
  \bottomrule
   \end{tabular}
        \vspace{-2mm}
    \end{minipage}
\end{table*}

\noindent \textbf{Speed Measurement.} Following the same protocol as \cite{vasu2023fastvit}, mobile latency is measured on iPhone 12 (iOS 16.5) using models exported by CoreML tools, while GPU throughput is measured with an NVIDIA RTX 4090 GPU.
For GPU throughput, we compile the traced model with TensorRT (v8.6.1.post1).
Unless otherwise specified, all methods are evaluated using their respective input sizes and a batch size of 1, following the aforementioned procedure.

\begin{table*}[t]
    \begin{minipage}{\textwidth}
    \centering
    \caption{
        \textbf{ImageNet-1k classification with Distillation.} 
        Comparison of SOTA efficient models on ImageNet-1K  classification, using DeiT~\cite{touvron2021deit} distillation recipe.
        }\label{tab:imagenet_distillation}
        \vspace{-1mm}
        \tablestyle{4pt}{0.8}
        \scriptsize
        \begin{tabular}{lcccccc}\toprule
  \multirow{3}[3]{*}{Model} & Image & Params & FLOPs & GPU & Mobile & Top-1 \\ 
  & Size &  &  & Throu. & Latency & Acc. \\
   & (Px) & (M) & (G) & \scalebox{0.9}{(imgs/s)} & (ms) & (\%) \\
  \midrule
  SwiftFormer-L1~\cite{shaker2023swiftformer} & $224^2$ & 12.1 & 1.6 & 943 & \textbf{1.6} & 80.9 \\
  EfficientFormerV2-S2~\cite{li2023efficientformerv2} & $224^2$ & 12.6 & 1.3 & 761 & 2.0 & 81.6 \\
  FastViT-SA12~\cite{vasu2023fastvit} & $256^2$ & 10.9 & 1.9 & 1274 & 2.0 & 81.9 \\
  \cellcolor[HTML]{efefef}\textbf{FFNet-1} & \cellcolor[HTML]{efefef}$256^2$ & \cellcolor[HTML]{efefef}13.7 & \cellcolor[HTML]{efefef}2.9 & \cellcolor[HTML]{efefef}\textbf{1430} & \cellcolor[HTML]{efefef}1.8 & \cellcolor[HTML]{efefef}\textbf{82.1} \\
  \midrule
  SwiftFormer-L3~\cite{shaker2023swiftformer} & $224^2$ & 28.5 & 4.0 & 609 & 3.0 & 83.0 \\
  EfficientFormerV2-L~\cite{li2023efficientformerv2} & $224^2$ & 26.1 & 2.6 & 598 & \textbf{2.8} & 83.3 \\
  FastViT-SA24~\cite{vasu2023fastvit} & $256^2$ & 20.6 & 3.8 & 596 & 3.0 & 83.4 \\
  \cellcolor[HTML]{efefef}\textbf{FFNet-2} & \cellcolor[HTML]{efefef}$256^2$ & \cellcolor[HTML]{efefef}26.9 & \cellcolor[HTML]{efefef}6.0 & \cellcolor[HTML]{efefef}\textbf{776} & \cellcolor[HTML]{efefef}3.1 & \cellcolor[HTML]{efefef}\textbf{83.7} \\
  \midrule
 \cellcolor[HTML]{efefef}\textbf{FFNet-3} & \cellcolor[HTML]{efefef}$256^2$ & \cellcolor[HTML]{efefef}48.3 & \cellcolor[HTML]{efefef}10.1 & \cellcolor[HTML]{efefef}\textbf{540} & \cellcolor[HTML]{efefef}\textbf{4.5} & \cellcolor[HTML]{efefef}\textbf{84.5} \\
  \bottomrule
   \end{tabular}
        \vspace{-1mm}
    \end{minipage}
\end{table*}

\subsection{FFNet for Efficient Image Recognition}
\noindent \textbf{ImageNet Classification.}
Following convention, FFNet is structured into four stages, each consisting of multiple FFNet blocks for hierarchical feature extraction.
The FFNet block is illustrated in Fig.~\ref{fig:FFNification} (c).
In the token mixer, for efficiency, we adopt a kernel size of 3$\times$3 in the first two stages and switch to 7$\times$7 in later stages.
In the channel mixer, we mainly use 7$\times$7 kernels, and the expansion ratio is set to 3 by default.
We evaluate FFNet on the ImageNet-1k~\cite{deng2009imagenet} dataset.
We follow the same training recipe in DeiT~\cite{touvron2021deit} for a fair comparison.
Specifically, we train our models for 300 epochs using AdamW optimizer with an initial learning rate of 1$\times 10^{-3}$ via cosine schedule and a weight decay of 0.05.
We set the total batch size as 1024 and the input size as $256^2$.
For $384^2$ resolution, we finetune the models for 30 epochs with weight decay 1$\times 10^{-8}$, learning rate of 5$\times 10^{-6}$, and batch size of 512.
Following DeiT, we also report the results of hard distillation using RegNetY-16GF~\cite{radosavovic2020regnet} as the teacher model.
Further details on the model specifications are provided in the Appendix.

\noindent\emph{\underline{Results.}}
In Tab.~\ref{tab:imagenet}, we compare our models against recent state-of-the-art methods.
Notably, the comparative results clearly show that our FFNet achieves the best speed-accuracy trade-off on two testbeds (GPU and mobile device).
In particular, our model demonstrates a substantial efficiency improvement over attention-based models~\cite{guo2022cmt, liu2021swin, si2022inceptionformer, tu2022maxvit, zhu2023biformer, ding2022davit, fan2024rmt, smt}, especially on the mobile device.
Furthermore, FFNet outperforms CNN-based models such as ConvNeXt, HorNet, and MogaNet, which use kernels of the same size, thereby affirming the superiority of the proposed MetaMixer framework.
FFNet also generalizes well to a larger image resolution and distillation-based training.
A thorough comparison with relevant methods is provided in the Appendix.

\begin{table}[t]
    \centering
    \caption{
        \textbf{COCO object detection and instance segmentation.}
        All statistics are calculated with image size (1280, 800).
        }\label{tab:cascademaskrcnn}
        \tablestyle{4.2pt}{0.8}
        \scriptsize
        \begin{tabular}{lcccc}
  \toprule 
        Backbone & GPU & Mobile &  $\text{AP}^{\text{box}}$  & $\text{AP}^{\text{mask}}$ \\ 
        & Throu.$\uparrow$ & Latency$\downarrow$ & (\%) & (\%)  \\ 
        \midrule
  SMT-S~\cite{smt} & 29 & OOM &51.9 &44.7\\
  Swin-S~\cite{liu2021swin} & 34 & OOM & 51.9 & 45.0 \\
  ConvNeXt-T~\cite{liu2022convnet} & 139 & 86.2 & 50.4 & 43.7 \\
  ConvNeXt-B~\cite{liu2022convnet} & 54 & 173.1 & 52.7 & 45.6 \\
  HorNet-S$_{7\times 7}$~\cite{rao2022hornet} & 66 & 124.9 & 52.7 & 45.6\\
  HorNet-B$_{7\times 7}$~\cite{rao2022hornet} & 44 & 182.2 & 53.3 & \textbf{46.1}\\
  RMT-S~\cite{fan2024rmt} &32 &- &53.2 &\textbf{46.1} \\
 \midrule
    \cellcolor[HTML]{efefef}FFNet-2 & \cellcolor[HTML]{efefef}141 & \cellcolor[HTML]{efefef}51.5 & \cellcolor[HTML]{efefef}51.8 & \cellcolor[HTML]{efefef}44.9 \\
    \cellcolor[HTML]{efefef}FFNet-3 & \cellcolor[HTML]{efefef}92 & \cellcolor[HTML]{efefef}80.5 & \cellcolor[HTML]{efefef}52.8 & \cellcolor[HTML]{efefef}45.6 \\
    \cellcolor[HTML]{efefef}FFNet-4 & \cellcolor[HTML]{efefef}62 & \cellcolor[HTML]{efefef}136.3 & \cellcolor[HTML]{efefef}\textbf{53.4} & \cellcolor[HTML]{efefef}45.9 \\
  \bottomrule
  \end{tabular}
        \vspace{-3mm}
\end{table}

\begin{table}[t]
     \begin{minipage}{\columnwidth}
    \centering
    \caption{
        \textbf{2D semantic segmentation.} For a thorough assessment of efficiency, we report backbone latency at the backbone level and total forward pipeline latency at the system level comparison. System-level latency is exceptionally measured on an RTX A6000 with PyTorch.
        Backbone statistics are calculated with image size (2048, 512).
        For Cityscapes, the input size is (2048, 1024). “SS”/“MS” denote single- and multi-scale testing, respectively.
        }\label{tab:2dsemseg}
        \tablestyle{3.0pt}{0.9}
        \scriptsize
          \begin{tabular}
  {lccccc}\toprule
  \multirow{3}[3]{*}{Method} & Param & FLOPs & GPU & Mobile & mIoU \\ 
   &  &  & Latency & Latency & SS / MS \\
    & (M) & (G) & (ms) $\downarrow$ & (ms) $\downarrow$  & (\%) \\
  \midrule
  \addlinespace[2pt]
  \multicolumn{6}{c}{\scalebox{0.95}{\emph{Backbone-level Comparison on ADE20K with UperNet 160K}}} \\
  \addlinespace[2pt]
  Swin-B~\cite{liu2021swin} & 121 & 1188 & 36 & OOM & 48.1 / 49.7 \\
  ConvNeXt-T~\cite{liu2022convnet} & 60 & 939 & 8 & 70 & 46.0 / 46.7 \\
  ConvNeXt-B~\cite{liu2022convnet} & 122 & 1170 & 23 & 170 & 49.1 / 49.9 \\
  SMT-S~\cite{smt} & 50 & 935 &35 &1198 &49.2 / 50.2 \\
  DaViT-B~\cite{ding2022davit} & 121 & 1175 &24 &OOM &49.4 /\hspace{2.4mm}-\hspace{2.3mm} \\
  RMT-S~\cite{fan2024rmt} &56 &937 &34 &- &49.8 /\hspace{2.4mm}-\hspace{2.3mm} \\
  HorNet-S$_{7\times 7}$~\cite{rao2022hornet} & 81 & 1030 & 17 & 112 & 49.2 / 49.8 \\
  HorNet-B$_{7\times 7}$~\cite{rao2022hornet} & 121 & 1174 & 26 & 169 & 50.0 / 50.5 \\
  HorNet-B$_{\rm GF}$~\cite{rao2022hornet} & 126 & 1171 & - & - & 50.5 / 50.9 \\
  MogaNet-B~\cite{li2024moganet} & 74 & 1050 &30 &310 & 50.1 /\hspace{2.4mm}-\hspace{2.3mm} \\
  FocalNet-B(LRF)~\cite{yang2022focalnet} & 126 & 1192 &23 &- &50.5 / 51.4 \\
  \midrule
  \cellcolor[HTML]{efefef}FFNet-2 & \cellcolor[HTML]{efefef}58 & \cellcolor[HTML]{efefef}942 & \cellcolor[HTML]{efefef}7 & \cellcolor[HTML]{efefef}45 & \cellcolor[HTML]{efefef}47.1 / \cellcolor[HTML]{efefef}47.8 \\
  \cellcolor[HTML]{efefef}FFNet-3 & \cellcolor[HTML]{efefef}80 & \cellcolor[HTML]{efefef}1010 & \cellcolor[HTML]{efefef}11 & \cellcolor[HTML]{efefef}69 & \cellcolor[HTML]{efefef}49.6 / \cellcolor[HTML]{efefef}50.2 \\
  \cellcolor[HTML]{efefef}FFNet-4 & \cellcolor[HTML]{efefef}113 & \cellcolor[HTML]{efefef}1158 & \cellcolor[HTML]{efefef}16 & \cellcolor[HTML]{efefef}124 & \cellcolor[HTML]{efefef}\textbf{50.7} / \cellcolor[HTML]{efefef}\textbf{51.7} \\
  \midrule
  \addlinespace[2pt]
  \multicolumn{6}{c}{\emph{System-level Comparison on ADE20K}} \\
  \addlinespace[2pt]
  SETR-MLA~\cite{zheng2021setr} & 310 & 368 & 82 & - & 47.5 / 49.4 \\ 
  Segformer-B3~\cite{xie2021segformer}& 47 & 79 & 256 & - & 49.4 / 50.0 \\
  MaskFormer~\cite{cheng2021maskformer}& 63 & 79 & - & - & 49.8 / 51.0 \\
  SegNeXt-B~\cite{guo2022segnext} & 28 & 35 & 127 & - & 48.5 / 49.9 \\
  EfficientViT-L1~\cite{liu2023efficientvit} & 40 & 36 & 80 & - & 49.2 /\hspace{2.3mm}-\hspace{2.3mm}  \\
  \cellcolor[HTML]{efefef}$\text{FFNet}_{\textit{seg}}$ & \cellcolor[HTML]{efefef}68 & \cellcolor[HTML]{efefef}74 & \cellcolor[HTML]{efefef}65 & \cellcolor[HTML]{efefef}- & \cellcolor[HTML]{efefef}\textbf{50.1} / \cellcolor[HTML]{efefef}\textbf{51.2} \\
  \midrule
  \addlinespace[2pt]
  \multicolumn{6}{c}{\emph{System-level Comparison on Cityscapes}} \\
  \addlinespace[2pt]
  Segformer-B3~\cite{xie2021segformer} & 47 &963 & 252 & - & 81.7 / 83.3 \\
  SegNeXt-L~\cite{guo2022segnext} & 49 & 578 & 307 & - & \textbf{83.2} / 83.9 \\
  \cellcolor[HTML]{efefef}$\text{FFNet}_{\textit{seg}}$ & \cellcolor[HTML]{efefef}68 & \cellcolor[HTML]{efefef}577 & \cellcolor[HTML]{efefef}224 & \cellcolor[HTML]{efefef}- & \cellcolor[HTML]{efefef}\textbf{83.2} / \cellcolor[HTML]{efefef}\textbf{84.1} \\
  \bottomrule
   \end{tabular}
   
        \vspace{-2mm} 
    \end{minipage}
\end{table}

\noindent\textbf{Object Detection and Instance Segmentation.}
We evaluate the performance of FFNet as the backbone of Cascade Mask R-CNN~\cite{cai2018cascade} on the COCO instance segmentation task~\cite{lin2014coco}.
We adopt standard training recipe~\cite{liu2021swin, liu2022convnet} using 3$\times$ schedule with multi-scale training.

\noindent\emph{\underline{Results.}} The results presented in Tab.~\ref{tab:cascademaskrcnn} indicate that FFNet models outperform baselines across all scales.
Remarkably, our models demonstrate impressive efficiency gains compared to the SOTA baselines.
At comparable performance levels, FFNets run nearly twice as fast as RMT-S~\cite{fan2024rmt} on the GPU and ConvNeXt-B on the iPhone 12.

\noindent\textbf{2D Semantic Segmentation.}
For comparison at the backbone level, we utilize ImageNet-pretrained models as the backbone of UperNet~\cite{xiao2018upernet}. Furthermore, to enable a system-level comparison, we propose $\text{FFNet}_{\textit{seg}}$, a model similar to the original FFNet but incorporating task-specific modifications.
$\text{FFNet}_{\textit{seg}}$ adopts the standard backbone-head (encoder-decoder) architecture. 
Similar to the backbone, the head also employs FFNet blocks, thereby avoiding complex operations such as global self-attention \cite{cheng2021maskformer} and matrix decomposition \cite{guo2022segnext}.
Following ~\cite{guo2022segnext}, we input the features from the last three stages into the head, where they are concatenated and processed through an FFN.
More details are presented in Implementation details section.

We evaluate the effectiveness of our FFNet on two representative benchmark datasets: ADE20K~\cite{zhou2017ade20k} and Cityscapes~\cite{cordts2016cityscapes}.
For backbone-level comparison, our models are trained for 160k-iterations with a batch size of 16, following standard recipe~\cite{liu2021swin, liu2022convnet}.
For system-level comparisons, we adopt the same training settings in SegNeXt~\cite{guo2022segnext} and use ImageNet-1k pretrained models for initialization.
We use MMSegmentation toolbox~\cite{mmseg2020} for all semantic segmentation experiments.

\noindent\emph{\underline{Results.}}
Tab.~\ref{tab:2dsemseg} shows that our methods excel not only as backbones but also as standalone segmentors, highlighting our model's superior performance-latency trade-off.
For instance, FFNet-4 achieves a $0.6$ mIoU improvement over MogaNet-B while being 1.9$\times$/2.5$\times$ faster on the GPU/mobile device, respectively.
In addition, $\text{FFNet}_{\textit{seg}}$ consistently outperforms task-specific prior arts.

\begin{table*}[t]
    \centering
    \caption{
        \textbf{Comparisons on representative benchmarks for SR models at $\times$4 scale.}
        Performance metrics are calculated on the \textbf{Y}-channel.
        All computational statistics are calculated corresponding to an HR image of size 1280$\times$720.
        Memory denotes peak GPU memory consumption during the inference phase.
        Best and second-best results are bolded and underlined, respectively.
        }\label{tab: super-resolution}
        \tablestyle{1pt}{1.0}
        \tiny
        \begin{tabular}{@{}l|ccc|c|ccccc@{}}
\toprule
Methods  &
  \multicolumn{3}{c|}{Latency (ms)} &
  \multirow{2}{*}{\rotatebox{90}{\scalebox{0.75}{Dataset}}} &
  \multicolumn{5}{c}{\begin{tabular}[c]{@{}c@{}}Performance (PSNR/SSIM) $\uparrow$ \end{tabular}} \\ \cmidrule(lr){2-4} \cmidrule(l){6-10}
 &
  \begin{tabular}[c]{@{}c@{}}\tiny RTX4090\end{tabular} &
  \begin{tabular}[c]{@{}c@{}}\tiny iPhone12\end{tabular} &
  \begin{tabular}[c]{@{}c@{}}\tiny M2Air\end{tabular} &
  &
  Set5 &
  Set14 &
  B100 &
  Urban100 &
  Manga109 \\ \midrule
$\text{ELAN-light}$ & 19 &OOM & \underline{521} & \multirow{6}{*}{\rotatebox{90}{DIV2K}} &  32.43/0.8975 & 28.78/0.7858 & 27.69/0.7406 & 26.54/0.7982 & 30.92/0.9150 \\
$\text{SwinIR-NG}$ & 249 & - & - & & 32.44/0.8980 & \underline{28.83}/0.7870 & \underline{27.73}/0.7418 & 26.61/0.8010 & 31.09/0.9161 \\
$\text{SRFormer-l}$ & 214 & OOM & OOM & & \underline{32.51}/\underline{0.8988} & 28.82/\underline{0.7872} & \underline{27.73}/\underline{0.7422} & \textbf{26.67}/\textbf{0.8032} & \underline{31.17}/\underline{0.9165} \\
$\text{MambaIR-l}$  & 52 & - & - & & 32.42/0.8977 & 28.74/0.7847 & 27.68/0.7400 & 26.52/0.7983 & 30.94/0.9135 \\
$\text{SMFANet}+$  & \underline{12} & OOM & 653 & & 32.43/0.8979 & 28.77/0.7849 & 27.70/0.7400 & 26.45/0.7943 & 31.06/0.9138 \\
\cellcolor[HTML]{efefef}\textbf{$\text{FFNet}_{\textit{sr}}$-l} & \cellcolor[HTML]{efefef}\textbf{10} & \cellcolor[HTML]{efefef}\textbf{110} & \cellcolor[HTML]{efefef}\textbf{100} &\cellcolor[HTML]{efefef} & \cellcolor[HTML]{efefef}\textbf{32.52}/\textbf{0.8993} & \cellcolor[HTML]{efefef}\textbf{28.88}/\textbf{0.7884} & \cellcolor[HTML]{efefef}\textbf{27.77}/\textbf{0.7424} & \cellcolor[HTML]{efefef}\underline{26.66}/\underline{0.8018} &  \cellcolor[HTML]{efefef}\textbf{31.32}/\textbf{0.9170} \\ \midrule
$\text{OmniSR}$   & 25 & -            & -            & \multirow{5}{*}{\rotatebox{90}{{DF2K}}} & 32.57/0.8993 & 28.95/\underline{0.7898} & 27.81/0.7439 & \textbf{26.95}/\textbf{0.8105} & 31.50/\underline{0.9192} \\
$\text{DAT-light}$ & 530 & OOM          & OOM          &                        & 32.57/0.8991 & 28.87/0.7879 & 27.74/0.7428 & 26.64/0.8033 & 31.37/0.9178 \\
$\text{SAFMN-l}$   & 19 & OOM          & \underline{218}         &                        & \underline{32.65}/\underline{0.9005} & \underline{28.96}/\underline{0.7898} & \underline{27.82}/\underline{0.7440} & 26.81/0.8058 & \underline{31.59}/\underline{0.9192} \\ 
$\text{SMFANet}+$   & \textbf{12} & OOM            & 653            & & 32.51/0.8985 & 28.87/0.7872 & 27.74/0.7412 & 26.56/0.7976 & 31.29/0.9163 \\ \cellcolor[HTML]{efefef}\textbf{$\text{FFNet}_{\textit{sr}}$}  & \cellcolor[HTML]{efefef}\underline{14} & \cellcolor[HTML]{efefef}\textbf{145} & \cellcolor[HTML]{efefef}\textbf{133} &        \cellcolor[HTML]{efefef}                & \cellcolor[HTML]{efefef}\textbf{32.69}/\textbf{0.9009} & \cellcolor[HTML]{efefef}\textbf{29.01}/\textbf{0.7910} & \cellcolor[HTML]{efefef}\textbf{27.84}/\textbf{0.7449} & \cellcolor[HTML]{efefef}\underline{26.93}/\underline{0.8095} & \cellcolor[HTML]{efefef}\textbf{31.73}/\textbf{0.9209} \\ \bottomrule
\end{tabular}
        \vspace{-2mm}
\end{table*}

\subsection{Generalizing FFNet beyond Recognition}

\noindent\textbf{Efficient Super-Resolution (SR).} In this part, we explore the effectiveness of FFNet beyond recognition, extending its evaluation to low-level vision tasks.
$\text{FFNet}_{\textit{sr}}$ consists of three parts: shallow feature extractor~(H$_{SF}$), deep feature extractor~(H$_{DF}$), and upscale module~(H$_{Rec}$) following residual-in-residual architecture~\cite{RCAN}.
H$_{SF}$ extracts shallow features from a Low-Resolution~(LR) image using a 3$\times$3 convolution.
The deep feature extractor H$_{DF}$, comprising a stack of FFNet blocks, refines shallow features iteratively.
We only use 3$\times$3 depthwise convolutions for spatial modeling, with 96 base channels and an FFN expansion ratio of 2.
$\text{FFNet}_{\textit{sr}}$-light and $\text{FFNet}_{\textit{sr}}$ employ 36 and 48 blocks, respectively. 
We employ an anchor-based long residual connection~\cite{ABPN} where the LR image is repeated RGB-wise and added at the end of the H$_{DF}$.
Finally, H$_{Rec}$ reconstructs the High-Resolution image by up-sampling refined features using pixelshuffle.

$\text{FFNet}_{\textit{sr}}$-light and $\text{FFNet}_{\textit{sr}}$ are trained on the DIV2K~\cite{EDSR} and DF2K (DIV2K + Flicker2K~\cite{Flicker2K}) dataset, respectively.
Five standard datasets are adopted for testing: Set5~\cite{Set5}, Set14~\cite{Set14}, BSD100~\cite{B100}, Urban100~\cite{Urban100}, and Manga109~\cite{Manga109}.
We use PSNR and SSIM to evaluate the SR performance on the Y channel from the YCbCr color space after cropping 4 pixels at boundary~\cite{VDSR}.
Our models are trained by minimizing L1 loss and frequency reconstruction loss~\cite{MAXIM} with a weight factor set of 0.05.
For more details, please refer to the Appendix.

\noindent\emph{\underline{Results.}}
In Tab.~\ref{tab: super-resolution}, we compare our models against recent state-of-the-art SR models~\cite{zhang2022elan, liang2021swinir, srformer, guo2024mambair, SMFANet, SAFMN, wang2023omnisr, chen2023dat}.
Notably, $\text{FFNet}_{\textit{sr}}$ significantly outperforms SR specialists in terms of efficiency, running seamlessly even on resource-constrained devices, demonstrating its feasibility for real-world applications.
$\text{FFNet}_{\textit{sr}}$, employing 3$\times$3 depthwise convolutions within the MetaMixer, not only captures high-frequency details effectively but also allows for the stacking of more blocks within the same computational budget compared to the previous SOTA models~\cite{SAFMN, chen2023dat, wang2023omnisr, lee2026partial, Lee_2025_ICCV} that use complex mixers, thereby enhancing its ability to capture long-range contexts.
\emph{The impressive results suggest that our mixer design is not only effective for high-level tasks such as recognition but also well-suited for solving low-level problems.}

Visual comparisons are provided in the Appendix.

\vspace{-1mm}
\subsection{Generalizing FFNet beyond Image}
\vspace{-0.5mm}
In this section, we introduce models that adapt FFNet for point cloud and time series data by simply substituting its 2D convolutions with 1D/3D convolutions.

\begin{table}[t]
    \begin{minipage}{\textwidth}
    \centering
    \caption{
        \textbf{3D semantic segmentation.}
        }\label{tab:3d_sem_seg}
        \tablestyle{4.2pt}{1.0}
        \scriptsize
        \begin{tabular}{lcccccc}
\toprule
3D Sem. Seg.&\multicolumn{2}{c}{ScanNet~\cite{dai2017scannet}} &\multicolumn{2}{c}{ScanNet200~\cite{rozenberszki2022language}} &\multicolumn{2}{c}{S3DIS~\cite{armeni2016s3dis}} \\\cmidrule(lr){2-3} \cmidrule(lr){4-5} \cmidrule(lr){6-7}
Model &Val &Test &Val &Test &Area5 &6-fold \\\midrule
PointNeXt &71.5 &71.2 &- &- &70.5 &74.9 \\
ST &74.3 &73.7 &- &- &72.0 &- \\
OctFormer &75.7 &\textbf{76.6} &\underline{32.6} &32.6 &- &- \\
Swin3D &\textbf{76.4} &- &- &- &\underline{72.5} &\underline{76.9} \\
OA-CNNs & \underline{76.1} & 75.6 & 32.3 & \underline{33.3} & 71.1 & - \\
PTv2 &75.4 &74.2 &30.2 &- &71.6 &73.5 \\
\cellcolor[HTML]{efefef}\textbf{$\text{FFNet}_{\textit{3D}}$} & \cellcolor[HTML]{efefef}\underline{76.1} & \cellcolor[HTML]{efefef}\underline{76.5} & \cellcolor[HTML]{efefef}\textbf{33.8} & \cellcolor[HTML]{efefef}\textbf{34.5} & \cellcolor[HTML]{efefef}\textbf{72.6} & \cellcolor[HTML]{efefef}\textbf{77.1} \\
\bottomrule
\end{tabular}
        \vspace{-2mm}
    \end{minipage}
\end{table}

\noindent\textbf{3D Semantic Segmentation.} Following~\cite{wu2022point}, we adopt the U-Net~\cite{ronneberger2015unet} structure, comprising four stages of encoders and decoders with block depths of [2, 2, 6, 2] and [1, 1, 1, 1], respectively.
We utilize Grid Pooling~\cite{wu2022point}, setting all grid size multipliers to $\times$2, which represents the expansion ratio over the preceding pooling stage.
For token mixing, we employ submanifold convolution~\cite{graham2018submanifold}. Additionally, to manage memory-constrained scenarios requiring small batch sizes, we use both batch normalization and pre-norm (Layer Norm) scheme.
Due to the implementation limitations of submanifold convolution, which does not support depthwise operation, we restrict the kernel size to 3$\times$3$\times$3 to handle overfitting issues. Despite these constraints, the obtained results are promising, demonstrating that even small kernels are highly effective within the MetaMixer framework and U-Net architecture.

We conduct experiments using our $\text{FFNet}_{\textit{3D}}$ on the representative datasets: ScanNetv2~\cite{dai2017scannet}, ScanNet200~\cite{rozenberszki2022language}, and S3DIS~\cite{armeni2016s3dis}.
We use Pointcep~\cite{pointcept2023} codebase specialized for point cloud perception tasks.
Detailed descriptions of architecture and training settings can be found in the Appendix.

\noindent\emph{\underline{Results.}} 
$\text{FFNet}_{\textit{3D}}$ outperforms the task-specialized competitors, as shown in Tab.~\ref{tab:3d_sem_seg}.
Notably, our architecture design is embarrassingly simple compared to the baselines~\cite{yang2023swin3d}, reinforcing the idea that the MetaMixer framework is inherently a key performance driver.
We expect that incorporating adaptive~\cite{peng2024oacnn}/large-kernel~\cite{chen2023largekernel3d} convolution into our mixer design could yield further improvements.

\begin{table*}[t]
    \begin{minipage}{\textwidth}
    \centering
    \caption{
        \textbf{Multivariate time series long-term forecasting results} with prediction lengths $S\in\{96, 192, 336, 720\}$ and fixed lookback length $T=96$. Results are averaged from all prediction lengths. ETT results further averaged by subsets. The baseline numbers are borrowed from iTransformer.
        }\label{tab:forecasting_result}
        \tablestyle{4.2pt}{1.0}
        \scriptsize
        \resizebox{0.8\linewidth}{!}{
  \begin{threeparttable}
  \renewcommand{\multirowsetup}{\centering}
  \setlength{\tabcolsep}{1.45pt}
  \begin{tabular}{c|cc|cc|cc|cc|cc}
    \toprule
    {\multirow{2}{*}{Models}} & 
    \multicolumn{2}{c}{\rotatebox{0}{\scalebox{0.75}{\textbf{$\text{FFNet}_{\textit{1D}}$}}}} &
    \multicolumn{2}{c}{\rotatebox{0}{\scalebox{0.8}{iTransformer}}} &
    \multicolumn{2}{c}{\rotatebox{0}{\scalebox{0.8}{PatchTST}}} &
    \multicolumn{2}{c}{\rotatebox{0}{\scalebox{0.8}{Crossformer}}} &
    \multicolumn{2}{c}{\rotatebox{0}{\scalebox{0.8}{{TimesNet}}}} \\
     &
    \multicolumn{2}{c}{\scalebox{0.8}{\textbf{(Ours)}}} &
    \multicolumn{2}{c}{\scalebox{0.8}{\cite{liu2024itransformer}}} & 
    \multicolumn{2}{c}{\scalebox{0.8}{\cite{PatchTST}}} & 
    \multicolumn{2}{c}{\scalebox{0.8}{\cite{Crossformer}}} & 
    \multicolumn{2}{c}{\scalebox{0.8}{\cite{Timesnet}}} \\
    \cmidrule(lr){2-3} \cmidrule(lr){4-5}\cmidrule(lr){6-7} \cmidrule(lr){8-9}\cmidrule(lr){10-11}
    {Metric}  & \scalebox{0.8}{MSE} & \scalebox{0.8}{MAE}  & \scalebox{0.8}{MSE} & \scalebox{0.8}{MAE}  & \scalebox{0.8}{MSE} & \scalebox{0.8}{MAE}  & \scalebox{0.8}{MSE} & \scalebox{0.8}{MAE}  & \scalebox{0.8}{MSE} & \scalebox{0.8}{MAE} \\
    \toprule
    
    \scalebox{0.95}{ETT} & \textbf{\scalebox{0.8}{0.376}} & \textbf{\scalebox{0.8}{0.390}} & {\scalebox{0.8}{{0.383}}} & {\scalebox{0.8}{0.399}} & \underline{\scalebox{0.8}{0.381}} & \underline{\scalebox{0.8}{0.397}} & \scalebox{0.8}{{0.685}} & \scalebox{0.8}{{0.578}} & \scalebox{0.8}{{0.391}} & \scalebox{0.8}{{0.404}}  \\
    \midrule 

    \scalebox{0.95}{ECL} & \underline{\scalebox{0.8}{0.182}} & \underline{\scalebox{0.8}{0.277}} & \textbf{\scalebox{0.8}{0.178}} &\textbf{\scalebox{0.8}{0.270}} & \scalebox{0.8}{0.205} & {\scalebox{0.8}{0.290}} & \scalebox{0.8}{0.244} & \scalebox{0.8}{0.334}  &{\scalebox{0.8}{0.192}} &\scalebox{0.8}{0.295}  \\
    \midrule

    \scalebox{0.95}{Exchange} & \textbf{\scalebox{0.8}{0.359}} & \textbf{\scalebox{0.8}{0.403}} &\underline{\scalebox{0.8}{0.360}} &\textbf{\scalebox{0.8}{0.403}} & \scalebox{0.8}{0.367} & \underline{\scalebox{0.8}{0.404}} & \scalebox{0.8}{0.940} & \scalebox{0.8}{0.707} &{\scalebox{0.8}{0.416}} &{\scalebox{0.8}{0.443}} \\

    \midrule 
    \scalebox{0.95}{Traffic} & {\scalebox{0.8}{0.494}} & {\scalebox{0.8}{0.328}} &\textbf{\scalebox{0.8}{0.428}} &\textbf{\scalebox{0.8}{0.282}} & \underline{\scalebox{0.8}{0.481}} & \underline{\scalebox{0.8}{0.304}} & \scalebox{0.8}{0.550} & \underline{\scalebox{0.8}{0.304}} &{\scalebox{0.8}{0.620}} &{\scalebox{0.8}{0.336}}  \\
    
    \midrule
    \scalebox{0.95}{Weather} & \textbf{\scalebox{0.8}{0.244}} & \textbf{\scalebox{0.8}{0.274}} &\underline{\scalebox{0.8}{0.258}} &\underline{\scalebox{0.8}{0.278}} & {\scalebox{0.8}{0.259}} & {\scalebox{0.8}{0.281}} & \scalebox{0.8}{0.259} & \scalebox{0.8}{0.315} &{\scalebox{0.8}{0.259}} &{\scalebox{0.8}{0.287}}  \\

    \bottomrule
  \end{tabular}
  \end{threeparttable}
}
        \vspace{-2mm}
    \end{minipage} 
\end{table*}

\noindent\textbf{Time Series Forecasting.} $\text{FFNet}_{\textit{1D}}$ employs a backbone-head structure with RevIN~\cite{kim2022reversible}, where the backbone consists of stacked FFNet blocks.
In our mixer design, we utilize a 51-sized kernel for 1D depthwise convolution to effectively catch up the global receptive fields of transformer- and MLP-based models.
In fact, directly applying the FFNet block to time series data is challenging due to the presence of variable dimension in addition to channel and temporal dimensions in multivariate setting.
To effectively capture both temporal and cross-variable dependencies, we modify FFN part using appropriate grouping and feature reshaping.
Detailed descriptions of these time series-specific modifications and training settings are available in the Appendix.
We conduct long-term forecasting experiments on popular 8 real-world datasets, including ETT (4 subsets), ECL, Exchange, Traffic, Weather.
Following \cite{liu2024itransformer}, we set the input length as 96 for all the results.
We compute the Mean Squared Error (MSE) and Mean Absolute Error (MAE) as metrics for multivariate time series forecasting. 

\emph{\underline{Results.}} 
Tab.~\ref{tab:forecasting_result} shows the competitive performance of $\text{FFNet}_{\textit{1D}}$ in multivariate time-series forecasting.
Remarkably, $\text{FFNet}_{\textit{1D}}$ achieves overall superior performance compared to the SOTA Transformer models, demonstrating the excellent task-generality of our convolutional mixer design.

\emph{
In summary, FFNets not only achieve strong performance but also substantially enhance efficiency across various devices, covering high-level recognition and low-level vision tasks.
Their seamless extension beyond image data underscores the MetaMixer as a crucial architectural foundation.
}

\begin{table}[t]
    \begin{minipage}{\textwidth}
    \centering
    \caption{
        \textbf{ImageNet 256$\times$256 generation performance with classifier-free guidance ($cfg = 1.5$).} Throughput is measured only for the U-Net, using an NVIDIA RTX A6000.
        }\label{tab:diffusion}
        \vspace{-2mm}
        \tablestyle{4.2pt}{1.0}
        \scriptsize
        \begin{tabular}{lcccc}
\toprule
Model & FLOPs &Throughput &FID$\downarrow$ &IS$\uparrow$ \\\midrule
U-DiT-B~\cite{tian2024udit} &22.2G &276.5 &3.10 &241.56 \\
\cellcolor[HTML]{efefef}U-FFNet-B &\cellcolor[HTML]{efefef}\textbf{19.6G} &\cellcolor[HTML]{efefef}\textbf{331.0} &\cellcolor[HTML]{efefef}\textbf{2.99} &\cellcolor[HTML]{efefef}\textbf{245.71} \\
\bottomrule
\end{tabular}
        \vspace{-2mm}
    \end{minipage}
\end{table}

\vspace{-0.5mm}
\subsection{Unified Mixer Design with MetaMixer}
Our MetaMixer framework, when combined with convolution, provides an efficient baseline while also opening a unified perspective that encompasses self-attention (see Appendix).
While convolutional interaction, such as FFNified attention, is more efficient than attention in high-resolution settings, the advantages of attention remain compelling—it excels in capturing high-level semantic representations and enables flexible operations through inter-token manner.
Furthermore, it is noted in~\cite{park2022howdovision} that convolution and self-attention complement each other, as convolution is adept at capturing high-frequency signals while self-attention specializes in low-frequency signals.
This implies that leveraging both worlds within the MetaMixer framework in a balanced manner can optimize both computational efficiency and representational power.

To demonstrate the efficacy of the aforementioned hybrid approach, we conduct ImageNet class-conditional generation experiments.
We choose U-DiT~\cite{tian2024udit} built upon Latent Diffusion~\cite{rombach2022stablediffusion} as our baseline and replace attention with FFNified attention in the encoder/decoder, except for the bottleneck stage.
FFNified attention efficiently extracts fine-grained details in high-resolution stages, while attention in the bottleneck stage models semantic information. As shown in Tab.~\ref{tab:diffusion}, this synergy enables our model to achieve improved FID-speed performance.
Samples generated using our model are presented in the Appendix.
\vspace{-0.5mm}
\section{Discussion and Conclusion}
\vspace{-0.5mm}
Although convolution has been extensively explored before, we are the first to integrate it into query-key-value interactions, introducing a versatile convolutional mixer design that offers general efficiency across various vision tasks and devices.
For a more detailed discussion on our relationship with previous CNNs, please refer to the Appendix.
Furthermore, by making effective use of large kernels, FFNified attention achieves an effective receptive field comparable to global self-attention (see Appendix).

In this work, we analyzed the FFN from a seldom-explored perspective and integrated self-attention and FFN-based mixers within a query-key-value framework.
We further abstracted this framework into a general mixer architecture, termed MetaMixer, where sub-operations are unspecified.
Based on the MetaMixer, we presented a convolutional mixer design with FFNified attention.
In numerous tasks where transformer-based and convolution-based models are in fierce competition, we merge both approaches in a completely new direction, and achieve a new state of the art with significant efficiency gains.
Moreover, the proposed unified mixer design provides deeper insights into the synergy between the two approaches.
These achievements underscore that the significance lies not in specific module per se, but in satisfying task-specific functionalities, starting with MetaMixer as the foundation.

\clearpage

\title{Supplementary Materials for \\
FFNet: MetaMixer-based Efficient Convolutional Mixer Design}

\titlerunning{FFNet}

\author{Seokju Yun \inst{1} \and
Dongheon Lee\inst{2} \and
Youngmin Ro\inst{2}
}
\authorrunning{S.~Yun et al.}

\institute{
KAIST AI
\and
University of Seoul \\
\url{https://github.com/ysj9909/FFNet}
}

\maketitle

\renewcommand{\thesection}{\Alph{section}}
\renewcommand{\thesubsection}{\thesection.\arabic{subsection}}
\renewcommand{\thesubsubsection}{\thesubsection.\arabic{subsubsection}}

\renewcommand{\thefigure}{S.\arabic{figure}}
\renewcommand{\thetable}{S.\arabic{table}}
\renewcommand{\theequation}{S.\arabic{equation}}


\setcounter{section}{0}
\setcounter{figure}{0}
\setcounter{table}{0}
\setcounter{equation}{0}

\makeatletter
\renewcommand{\theHsection}{supp.\Alph{section}}
\renewcommand{\theHsubsection}{\theHsection.\arabic{subsection}}
\renewcommand{\theHsubsubsection}{\theHsubsection.\arabic{subsubsection}}
\makeatother

\section{FFNs Are Key-Value  Memories} \label{app_sec: ffn_kvm}
In this section, we conduct an in-depth analysis of the Feed-Forward Network (FFN) within the trained vision model PoolFormer-M36~\cite{yu2021metaformer}. We identify the most activated keys on average for each class using the ImageNet-1K validation set. Surprisingly, keys specialized for each class demonstrate a human-interpretable correlation in semantically consistent regions, following the point-wise mechanism of the FFN (see Fig.~\ref{app_fig:FFNs_are_kvm} (a)). Moreover, these keys operate not in isolation but in a hierarchical and composite manner. For instance, certain keys are associated with higher-level concepts shared across classes, such as animal species and machine parts (see Fig.~\ref{app_fig:FFNs_are_kvm} (b)). Furthermore, we discover keys that strongly correlate with approximately 70\% to 85\% of the classes. For example, $K_{10}$ correlates with the surroundings of the target object, aiding in shape recognition (see Fig.~\ref{app_fig:special_key} (a)), while $K_{1395}$ correlates with a wide range of objects, serving as a detector of objectness (see Fig.~\ref{app_fig:special_key} (b)).

We also find that $K_{548}$ and $K_{1461}$ from the smaller variant PoolFormer-S12, as well as $K_{382}$ and $K_{2869}$ from the ViT-B trained on ImageNet-21K, operate similarly to the special keys discussed above. Additionally, ~\cite{li2022lazy} found that the activations of the FFN in ViT models exhibit sparsity, which aligns with the observations in Fig. 2 (a) of the main paper.
The above results suggest that FFNs generally operate as key-value memories across various models.

Interestingly, as shown in Fig.~\ref{app_fig:vit_kvm}, class-specific keys in ViT-B not only correlate locally with the target object but also with its immediate surroundings and low-informative background areas. This could be related to unexpected behaviors observed in attention layers, such as artifacts in attention maps~\cite{darcet2024register} and query-irrelevant attentions~\cite{ma2022close}.
Consequently, similar to previous work~\cite{kobayashi2024analyzing} that analyzed internal processing of FFNs in BERT and GPT-2 through attention maps, investigating FFN's key-value mechanism in relation to attention layers in vision models is also a promising future research direction.

\begin{figure}[t]
  \centering
  \includegraphics[width=0.8\columnwidth]{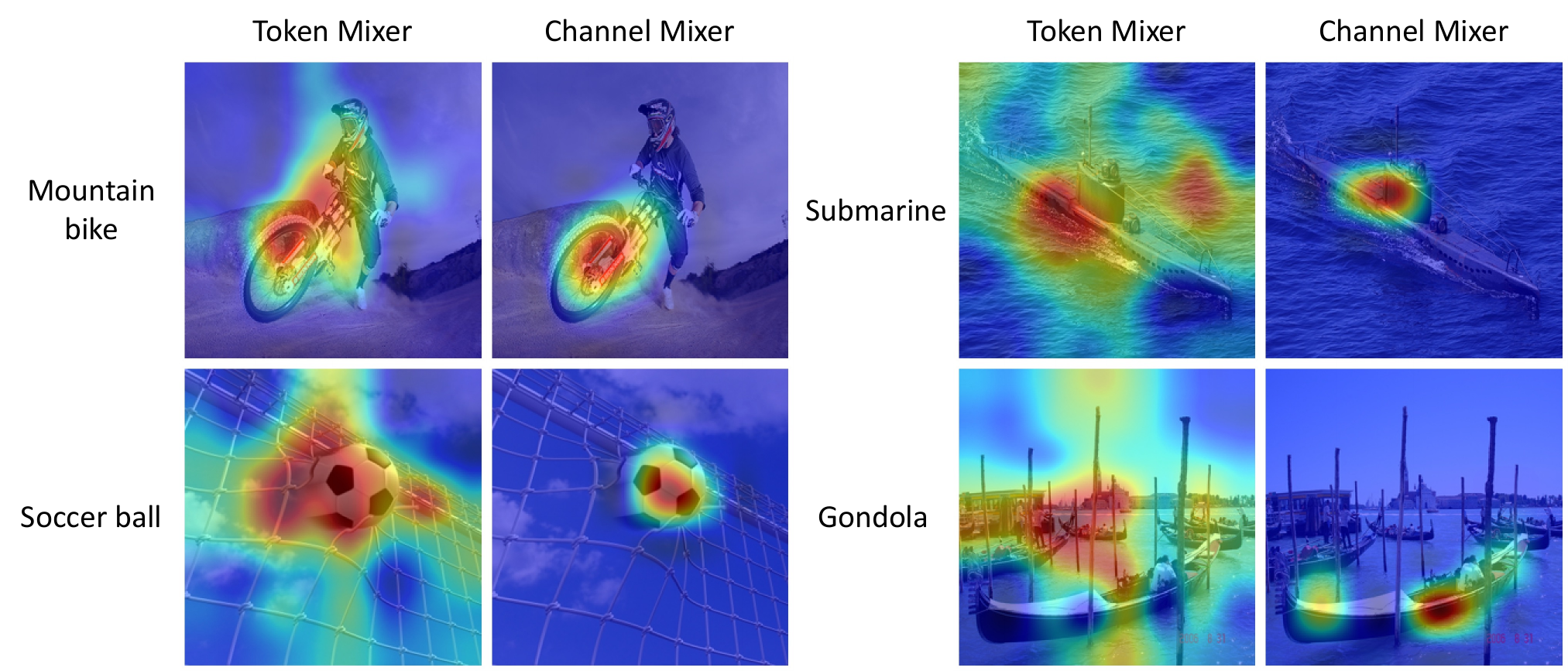}
  \caption{\small Examples of coefficient maps corresponding to the most activated key from the mixers in the final block of FFNet-3.
  }
  \label{app_fig:FFNet_are_kvm} 
\end{figure}

\begin{figure}[!t]
  \centering
  \includegraphics[width=0.9\columnwidth]{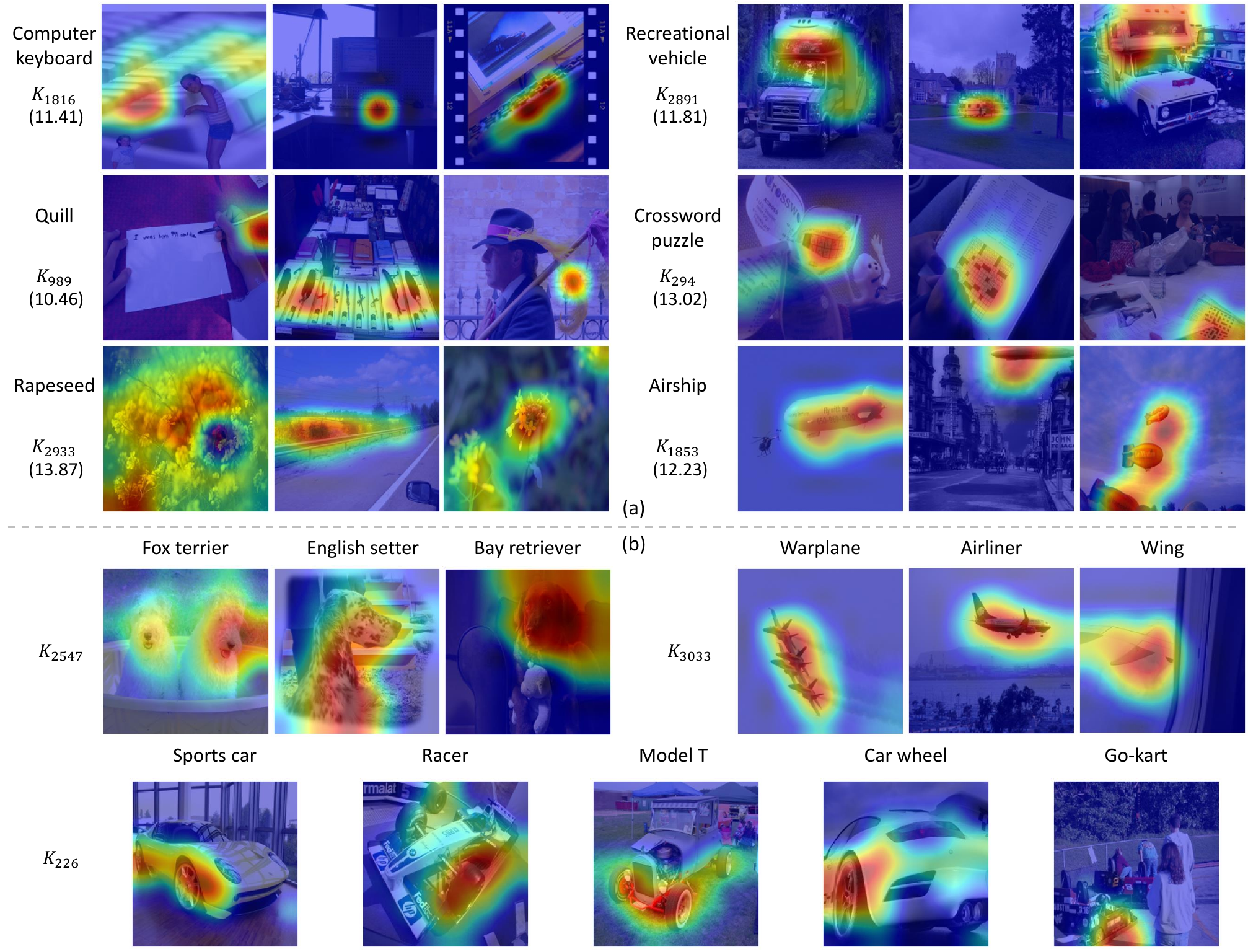}
  \caption{\small Visualization of the FFN's coefficient map in the last layer of PoolFormer-M36~\cite{yu2021metaformer}.
  (a) We visualize coefficient maps corresponding to the most activated keys for each class, where class-specific keys consistently correlate with decisive locations.
  The numbers in parentheses indicate the average values of the coefficients.
  (b) We identify keys that detect concepts shared across classes, such as animal species and machine parts, revealing that the keys capture patterns in the input through a hierarchical manner.
  Specifically, $K_{226}$ is mostly inactive except in classes that include wheels.
  }
  \label{app_fig:FFNs_are_kvm} 
\end{figure}

\begin{figure}[t]
  \centering
  \includegraphics[width=0.8\columnwidth]{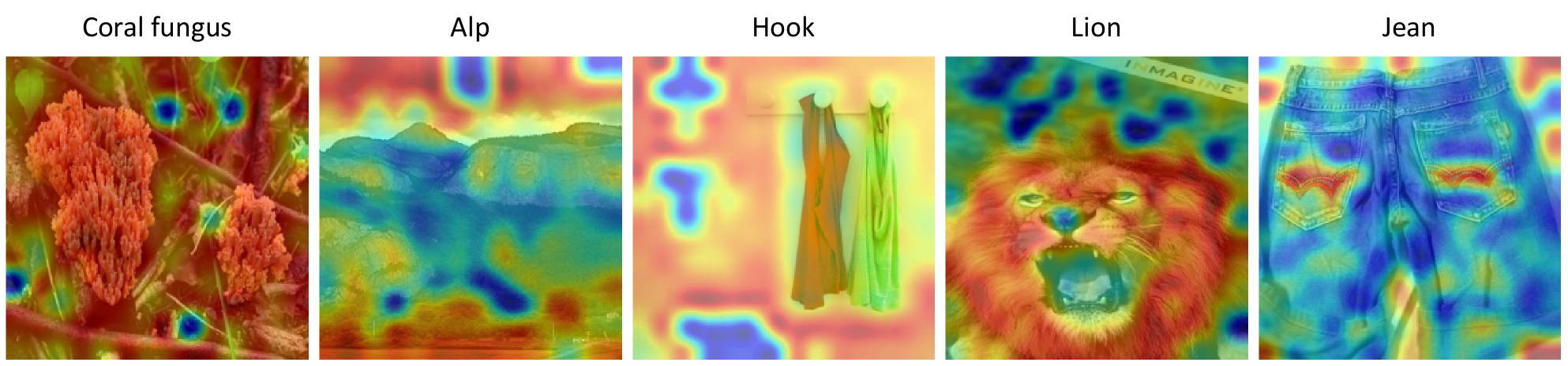}
  \caption{\small Coefficient maps from ViT-B~\cite{dosovitskiy2021vit} trained on ImageNet-21K. These examples exhibit behavioral patterns distinctly different from those observed in PoolFormer models, indicating a need for analysis alongside preceding attention layers. Detailed discussion is beyond the scope of this paper and is therefore omitted.
  }
  \label{app_fig:vit_kvm} 
\end{figure}

\begin{figure}[!t]
  \centering
  \includegraphics[width=0.85\columnwidth]{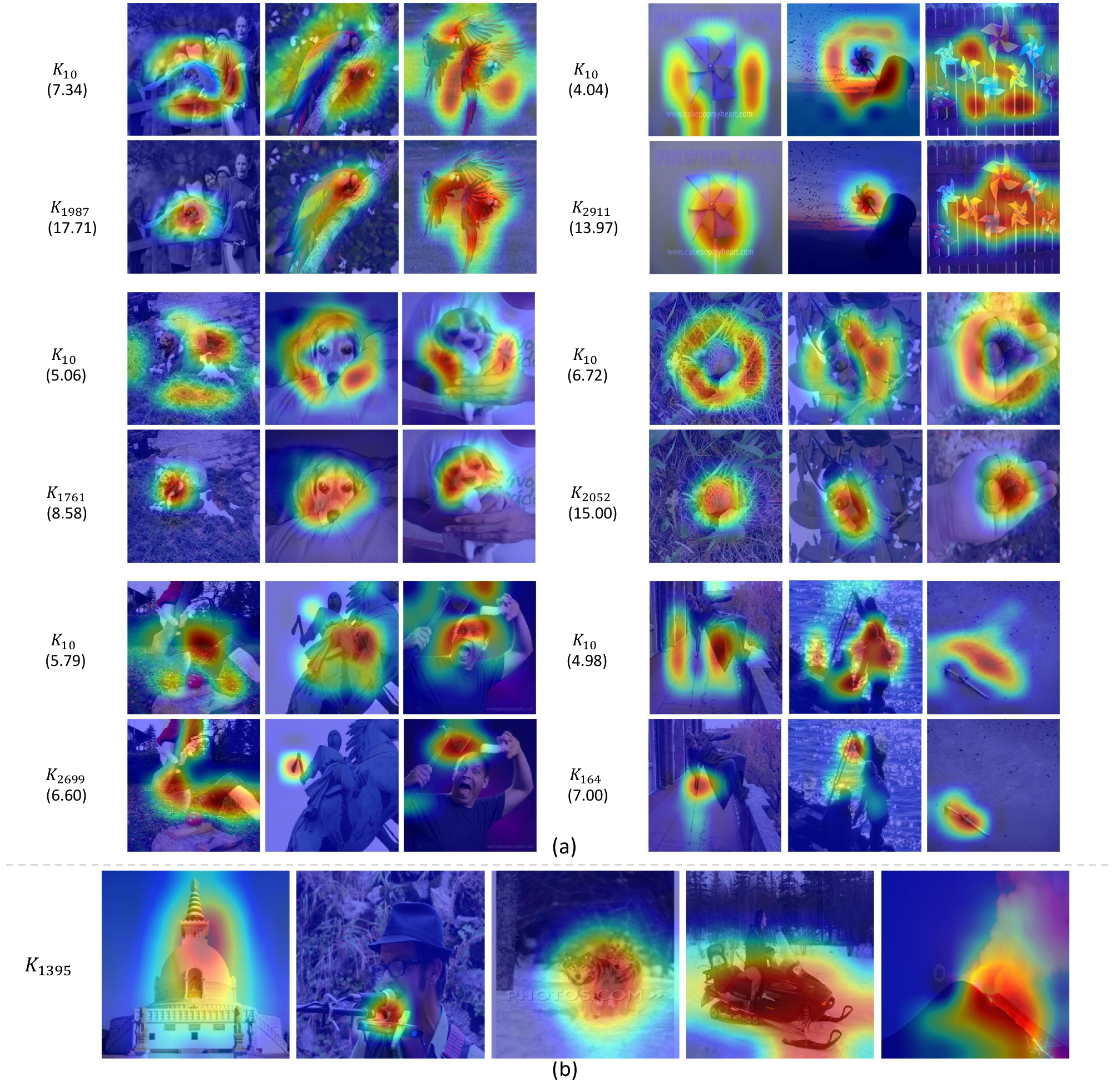}
  \caption{\small We analyze two representative class-agnostic keys in the last layer of PoolFormer-M36. $K_{10}$ and $K_{1395}$ are included in the most activated top-5 keys at rates of 71.7\% and 84.6\%, respectively. (a) $K_{10}$ correlates with the surrounding context of the target object, capturing its shape and aiding in context comprehension.
  The numbers in parentheses indicate the average values of the coefficients.
  (b) $K_{1395}$ consistently correlates across various types of objects.
  }
  \label{app_fig:special_key} 
\end{figure}

In Fig.~\ref{app_fig:FFNet_are_kvm}, we also present the coefficient maps of FFNet-3. Within the token mixer, the keys correlate with both the target object and its surrounding context, whereas in the channel mixer, the focus is centralized on the target object.
This demonstrates a complementary mechanism where the token mixer uses a large kernel to aggregate surrounding context, followed by the channel mixer concentrating on critical regions.

\begin{figure*}[t]
  \centering
  \includegraphics[width=\columnwidth]{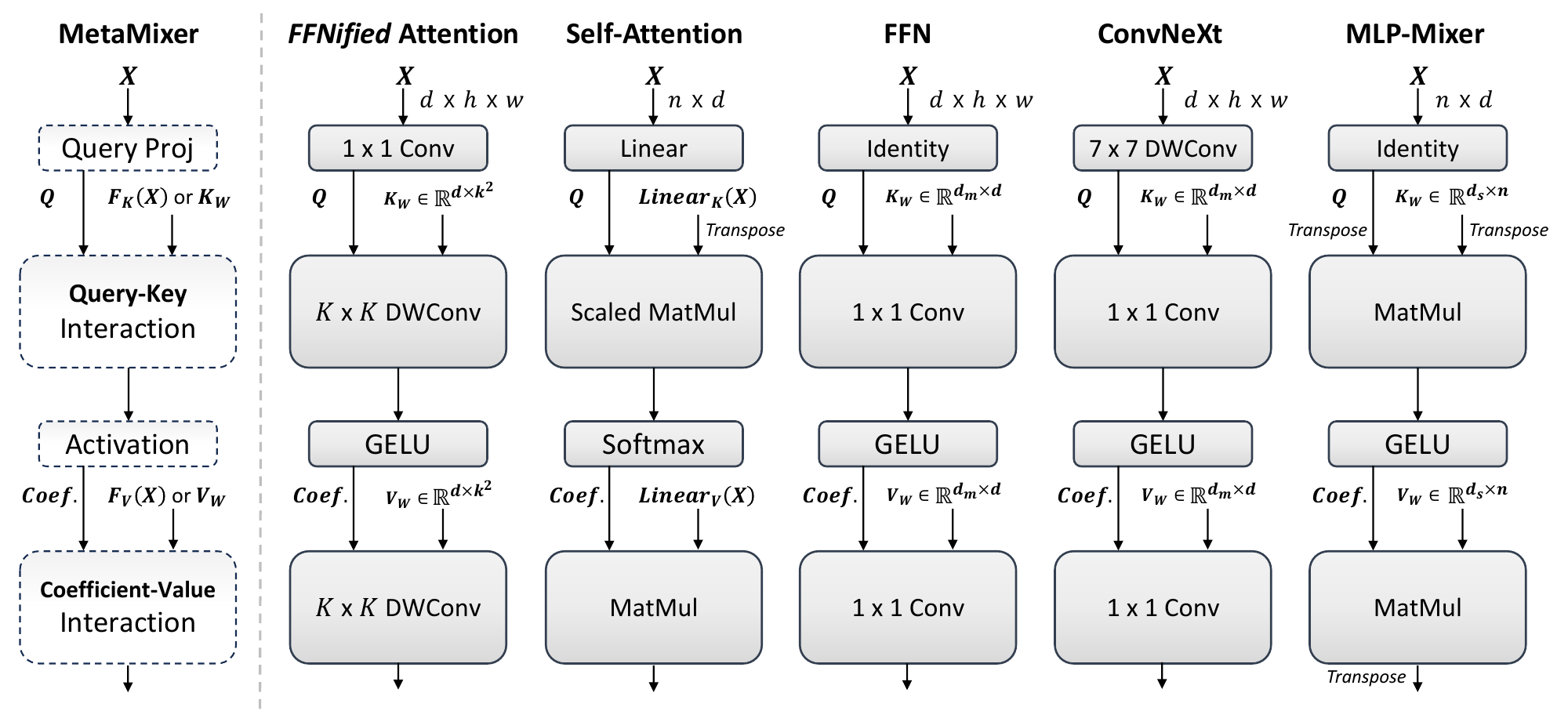}
  \caption{Generality of the MetaMixer. The notation for each mixer follows that introduced in its respective seminal paper. The variables $d$, $d_m$, $d_s$, and $n$ represent the base channel dimension, hidden dimension, hidden dimension of the spatial MLP, and token length (resolution), respectively.
  For ease of comparison, normalization layers are omitted. The MetaMixer framework encompasses the most successful mixers, demonstrating its modular nature.}
  \label{app_fig:metamixer_generality} 
\end{figure*}

\section{MetaMixer Generality} \label{app_sec: metamixer_generality}
As shown in Fig.~\ref{app_fig:metamixer_generality}, the most commonly used mixers are instantiated by specifying suboperations of the MetaMixer, underscoring that the query-key-value mechanism is a essential requirement for competitive performance. 
Specifically, the spatial MLP\footnote{Both MLP (Multi-Layer Perceptron) and FFN (Feed-Forward Network) generally refer to modules containing two fully-connected layers with an activation function between them. In this paper, we refer to this module as FFN overall, but use MLP specifically in the context of MLP-Mixer~\cite{tolstikhin2021mlpmixer}.} in MLP-Mixer~\cite{tolstikhin2021mlpmixer} operates via the same query-key-value mechanism as an FFN, but applies it along different axes—token and channel, respectively.
The ConvNeXt~\cite{liu2022convnet} block can be interpreted as employing a 7$\times$7 depthwise convolution as the query projection layer within an FFN.
Additionally, self-attention and FFNified attention utilize inter-token matrix multiplication and depthwise convolution, respectively, to compute query-key-value interactions.

\begin{table*}[t]
        \centering
        \caption{
        Comparison of inference speeds across various hardware platforms. “bs” denotes batch size. For CPU benchmarks, models are converted to ONNX runtime format and evaluated on an Intel(R) Xeon(R) Gold 5218R CPU @ 2.10GHz processor using a single thread. For Mobile and M2Air, we export the models using CoreML tools and report the median latency over 1,000 runs.
        }\label{app_tab:speed} 
        \tablestyle{2.5pt}{1.0}
        \tiny
        \begin{tabular}{lcccccccc}
  \toprule 
        Model & Type & Top-1 & \multicolumn{2}{c}{GPU Thro. (imgs/sec)} & \multicolumn{2}{c}{CPU Thro. (imgs/sec)} & Mobile & M2Air \\ \cmidrule(lr){4-5} \cmidrule(lr){6-7}
        & & (\%) & \hspace{3.5mm}bs 1 & \hspace{3.5mm}bs 64 & \hspace{3.5mm}bs 1 & \hspace{3.5mm}bs 16 &  Latency (ms) & Latency (ms)   \\ 
        \midrule
         MogaNet-T & CNN & 80.0 & \hspace{3.5mm}463 & \hspace{3.5mm}2169 & \hspace{3.5mm}25 & \hspace{3.5mm}72 &2.5 &2.0 \\
  \cellcolor[HTML]{efefef}FFNet-1 &\cellcolor[HTML]{efefef}CNN & \cellcolor[HTML]{efefef}81.3 & \cellcolor[HTML]{efefef}\hspace{3.5mm}1430 & \cellcolor[HTML]{efefef}\hspace{3.5mm}3965 & \cellcolor[HTML]{efefef}\hspace{3.5mm}38 & \cellcolor[HTML]{efefef}\hspace{3.5mm}62 & \cellcolor[HTML]{efefef}1.8 & \cellcolor[HTML]{efefef}1.3 \\      
  Swin-T  & ViT & 81.3 & \hspace{3.5mm}- & \hspace{3.5mm}2324 & \hspace{3.5mm}30 & \hspace{3.5mm}35 & 24.8 & 6.9 \\
  BiFormer-T  & ViT & 81.4 & \hspace{3.5mm}333 & \hspace{3.5mm}548 & \hspace{3.5mm}4 & \hspace{3.5mm}5 &- &- \\
  SMT-T & ViT & 82.2 & \hspace{3.5mm}786 & \hspace{3.5mm}2362 & \hspace{3.5mm}22 & \hspace{3.5mm}25 & 23.2 & 15.9 \\
  RMT-T  & ViT & 82.4 & \hspace{3.5mm}399 & \hspace{3.5mm}1569 & \hspace{3.5mm}- & \hspace{3.5mm}- &- &- \\
  DaViT-T  & ViT & 82.8 & \hspace{3.5mm}312 & \hspace{3.5mm}2120 & \hspace{3.5mm}23 & \hspace{3.5mm}25 &12.7 &8.3 \\
  ConvNeXt-T & CNN & 82.1 & \hspace{3.5mm}952 & \hspace{3.5mm}2633 & \hspace{3.5mm}41 & \hspace{3.5mm}50 & 3.7 & 2.4 \\
  \cellcolor[HTML]{efefef}FFNet-2 &\cellcolor[HTML]{efefef}CNN & \cellcolor[HTML]{efefef}\textbf{82.9} & \cellcolor[HTML]{efefef}\hspace{3.5mm}776 & \cellcolor[HTML]{efefef}\hspace{3.5mm}2163 & \cellcolor[HTML]{efefef}\hspace{3.5mm}31 & \cellcolor[HTML]{efefef}\hspace{3.5mm}52 & \cellcolor[HTML]{efefef}3.1 & \cellcolor[HTML]{efefef}2.1 \\
  \midrule
  iFormer-S & ViT & 83.4 & \hspace{3.5mm}500 & \hspace{3.5mm}1418 & \hspace{3.5mm}23 & \hspace{3.5mm}34 & 27.3 & 12.8 \\
  CMT-S & ViT & 83.5 & \hspace{3.5mm}407 & \hspace{3.5mm}1310 & \hspace{3.5mm}13 & \hspace{3.5mm}23 & 11.3 & 8.1 \\
  Swin-B & ViT & 83.5 & \hspace{3.5mm}- & \hspace{3.5mm}841 & \hspace{3.5mm}13 & \hspace{3.5mm}14 & 33.5 & 12.1 \\
  MaxViT-T & ViT & 83.6 & \hspace{3.5mm}- & \hspace{3.5mm}894 & \hspace{3.5mm}17 & \hspace{3.5mm}21 &120.1 &49.9 \\
  GC ViT-T2 & ViT & 83.7 & \hspace{3.5mm}- & \hspace{3.5mm}- & \hspace{3.5mm}18 & \hspace{3.5mm}25 &201.4 &69.1 \\
  SMT-S & ViT & 83.7 & \hspace{3.5mm}421 & \hspace{3.5mm}1289 & \hspace{3.5mm}12 & \hspace{3.5mm}13 & 41.5 & 28.3 \\
  BiFormer-S  & ViT & 83.8 & \hspace{3.5mm}162 & \hspace{3.5mm}257 & \hspace{3.5mm}2 & \hspace{3.5mm}3 &- &- \\
  MogaNet-S & CNN & 83.4 & \hspace{3.5mm}417 & \hspace{3.5mm}1300 & \hspace{3.5mm}21 & \hspace{3.5mm}47 & 4.3 & 3.1 \\
  HorNet-S$_{7\times 7}$& CNN & 83.8 & \hspace{3.5mm}351 & \hspace{3.5mm}954 & \hspace{3.5mm}12 & \hspace{3.5mm}18 & 5.4 & 3.7 \\
  ConvNeXt-B & CNN & 83.8 & \hspace{3.5mm}399 & \hspace{3.5mm}1263 & \hspace{3.5mm}16 & \hspace{3.5mm}17 & 8.1 & 5.6 \\
  FocalNet-B (LRF) & CNN & 83.9 & \hspace{3.5mm}287 & \hspace{3.5mm}828 & \hspace{3.5mm}12 & \hspace{3.5mm}31 &10.0 &7.0 \\
  \cellcolor[HTML]{efefef}FFNet-3 & \cellcolor[HTML]{efefef}CNN & \cellcolor[HTML]{efefef}\textbf{83.9} & \cellcolor[HTML]{efefef}\hspace{3.5mm}490 & \cellcolor[HTML]{efefef}\hspace{3.5mm}1431 & \cellcolor[HTML]{efefef}\hspace{3.5mm}22 & \cellcolor[HTML]{efefef}\hspace{3.5mm}40 & \cellcolor[HTML]{efefef}4.5 & \cellcolor[HTML]{efefef}2.9 \\
  \midrule
  CMT-B$_{r256}$ & ViT & 84.5 & \hspace{3.5mm}294 & \hspace{3.5mm}607 & \hspace{3.5mm}6 & \hspace{3.5mm}10 &13.2 &9.9 \\
  iFormer-S$_{r384}$ & ViT & \textbf{84.6} & \hspace{3.5mm}334 & \hspace{3.5mm}403 & \hspace{3.5mm}9 & \hspace{3.5mm}10 & 80.3 & 34.8 \\
  \cellcolor[HTML]{efefef}FFNet-3$_{r384}$ & \cellcolor[HTML]{efefef}CNN & \cellcolor[HTML]{efefef}84.5 & \cellcolor[HTML]{efefef}\hspace{3.5mm}387 & \cellcolor[HTML]{efefef}\hspace{3.5mm}595 & \cellcolor[HTML]{efefef}\hspace{3.5mm}12 & \cellcolor[HTML]{efefef}\hspace{3.5mm}20 & \cellcolor[HTML]{efefef}9.1 & \cellcolor[HTML]{efefef}7.1 \\
  \midrule
  MaxViT-T$_{r384}$ & ViT & 85.2 & \hspace{3.5mm}149 & \hspace{3.5mm}314 & \hspace{3.5mm}6 & \hspace{3.5mm}6 &247.6 &91.6 \\
  NFNet-F2$_{r384}$ & CNN & 85.1 & \hspace{3.5mm}194 & \hspace{3.5mm}347 & \hspace{3.5mm}6 & \hspace{3.5mm}10 &27.0 &18.0 \\
  ConvNeXt-B$_{r384}$ & CNN & 85.1 & \hspace{3.5mm}229 & \hspace{3.5mm}268 & \hspace{3.5mm}5 & \hspace{3.5mm}6 &21.3 &15.4 \\
  HorNet-B$_{7\times 7, r384}$& CNN & \textbf{85.3} & \hspace{3.5mm}175 & \hspace{3.5mm}233 & \hspace{3.5mm}5 & \hspace{3.5mm}6 & 21.3 & 16.3 \\
  \cellcolor[HTML]{efefef}FFNet-4$_{r384}$ & \cellcolor[HTML]{efefef}CNN & \cellcolor[HTML]{efefef}\textbf{85.3} & \cellcolor[HTML]{efefef}\hspace{3.5mm}282 & \cellcolor[HTML]{efefef}\hspace{3.5mm}394 & \cellcolor[HTML]{efefef}\hspace{3.5mm}10 & \cellcolor[HTML]{efefef}\hspace{3.5mm}13 & \cellcolor[HTML]{efefef}15.2 & \cellcolor[HTML]{efefef}10.8 \\
  \bottomrule
  \end{tabular}
\end{table*}

\section{Detailed Comparisons with Related Work} \label{app_sec: comparison}
\textbf{MLP-Mixer.} 
The term “FFNification” might initially evoke thoughts of spatial MLPs~\cite{tolstikhin2021mlpmixer} rather than our proposed \emph{FFNified attention} due to its direct implications.
However, our use of the term encompasses not only the adaptation of mechanisms but also aims to achieve the generality and efficiency characteristic of FFNs.
Spatial MLPs, on the other hand, suffer from parameter inefficiency~\cite{tolstikhin2021mlpmixer, zhao2021battle} and are impractical for high-resolution inputs, similar to the global self-attention.
They are also unable to handle multiple resolutions, limiting their applicability to downstream tasks.
Consequently, depthwise convolution emerges as an efficient and effective alternative, addressing these issues.
Unlike spatial MLPs, which share the same kernel across all channels, depthwise convolution applies different kernels to each channel.
To bridge the receptive field gap between convolution and spatial MLP, we employ large kernels (e.g., 7$\times$7 or 9$\times$9 for image recognition and 51 for time series).

\noindent\textbf{ConvNeXt.}
Both the ConvNeXt~\cite{liu2022convnet} block and FFNified attention mixer share basic elements such as convolution and GELU activation, yet they differ in the operating axes in the query-key-value mechanism.
FFNified attention actively extracts spatial features through two consecutive large-kernel convolutions, calculating query-key and coefficient-value interactions.
In contrast, the ConvNeXt block uses depthwise convolution as a means of query projection, insufficiently utilizing the surrounding context during channel interaction.
Hence, employing our proposed FFNified attention and ConvNeXt block as token mixer and channel mixer, respectively, provides a compact mixer design space.
In FFNified attention, it is natural to not expand channels, leading our FFNet to maintain a larger base channel size compared to similar-sized baseline models and thus reduce the expansion ratio in the channel mixer to 3.
This decision aligns with prior research findings~\cite{liu2023efficientvit, fasternet} that the expanded features of channel mixer contain substantial redundancy.
Thanks to our comprehensive design strategy and compact mixer space, our models demonstrate a better performance-speed trade-off than the ConvNeXt models especially on resource-constrained devices. (see Tab.~\ref{app_tab:speed}).

\noindent\textbf{Vision Transformer (ViT).}
Self-attention operates at the token level, employing matrix multiplication to calculate inter-token similarity and aggregate values using the resulting coefficients.
In contrast, FFNified attention operates at the channel level, detecting local patterns through depthwise convolution and applying a subsequent convolution to the resulting coefficients to aggregate values.
To reduce parameter complexity, self-attention shares attention weights across channels, while FFNified attention shares them across positions.
Another key difference lies in the key-value generation process: self-attention utilizes instance-specific key-value pairs through linear projections, while FFNified attention initializes key-values as static weights and updates them on the training set.

\noindent\textbf{Efficient ViTs.} Vanilla self-attention, being a token-based module, faces limitations in vision tasks with long token lengths (high-resolution tasks)~\cite{yun2024shvit}.
Swin transformer~\cite{liu2021swin} addresses this by performing self-attention within local windows.
However, window attention necessitates complex reshape operations, hindering efficient deployment across various inference platforms~\cite{liu2023efficientvit, yun2024shvit}.
Another line of work introduces sparsity into attention~\cite{zhu2023biformer, fan2024rmt, ding2022davit, tu2022maxvit, yun2023dynamic} or leveraging token merging layer (e.g., strided convolution or pooling) to shorten sequences before applying attention~\cite{guo2022cmt, si2022inceptionformer}.
Yet, these methods demand meticulous architectural tuning and generally lack efficiency on devices other than GPUs.
Conversely, FFNified attention, composed solely of hardware-friendly operations like convolution and GELU, exhibits high speed across diverse devices (see Tab.~\ref{app_tab:speed}).

\noindent\textbf{CNNs.} HorNet~\cite{rao2022hornet}, FocalNet~\cite{yang2022focalnet}, and MogaNet~\cite{li2024moganet} highlight the importance of multi-order interactions and propose high-order convolutional interaction modules.
SMT~\cite{smt} employs multi-head mixed convolution to efficiently extract multi-scale representations.
In contrast, by compactly utilizing convolution within the more fundamental \textbf{query-key-value framework (i.e. MetaMixer)}, rather than multi-order or multi-scale interactions, our models achieve an optimal performance-speed trade-off (see Tab.~\ref{app_tab:speed}).
On the other hand, large-kernel CNNs~\cite{ding2023unireplknet, ding2022scaling, liu2023more51,Lee_2025_ICCV, lee2026partial} depend on specialized libraries like GEMM, which hinders their general deployment across diverse environments such as TensorRT and CoreML.

\section{Effective Receptive Field Analysis}
\label{app_sec: erf}
Previous works~\cite{ding2022scaling, liu2023more51} have identified the Effective Receptive Field (ERF) as a key factor in enhancing CNN performance.
In this section, we mainly compare the ERFs of ConvNeXt and Swin Transformer, which use the same kernel (window) size, with our FFNet.
Following RepLKNet~\cite{ding2022scaling}, we sample and resize 50 images from the ImageNet validation set to 1024 $\times$ 1024, and measure the contribution of pixels on the input images to the central point of the feature map produced by the last layer. 
Contributions are aggregated and mapped onto a 1024 $\times$ 1024 matrix. 
We further quantify the ERF of each model by calculating the high-contribution area ratio $r$ of a minimum rectangle that encompasses the contribution scores over a certain threshold $t$. 

\begin{figure*}[t]
  \centering
  \begin{minipage}[c]{\columnwidth}
      \begin{minipage}[c]{0.6\columnwidth}
        \centering 
        \includegraphics[width=\columnwidth]{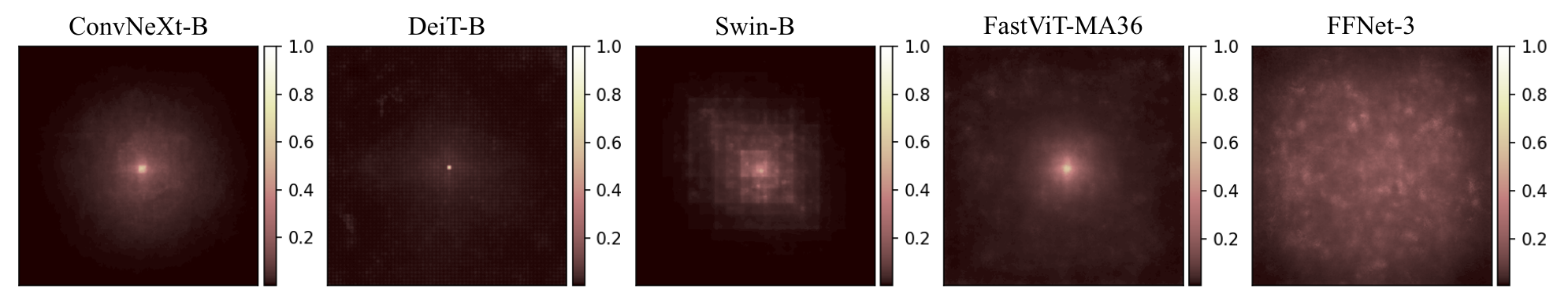}
      \end{minipage}
      \hfill
      \begin{minipage}[c]{0.38\columnwidth}
        \centering
        \resizebox{0.90\textwidth}{!}{%
            \begin{tabular}{@{}lcccc@{}}
            \toprule
            Model & t=20\% & t=30\% & t=50\% & t=99\% \\ \midrule
            ConvNeXt-B & 2.2\% & 3.7\% & 7.9\% & 64.6\% \\
            Swin-B & 1.0\% & 1.4\% & 3.6\% & 41.4\% \\
            FastViT-MA36 & 1.6\% & 3.2\% & 10.0\% & 97.5\% \\
            DeiT-B & 3.7\% & 8.2\% & 24.5\% & 98.3\% \\
            \cellcolor[HTML]{efefef}FFNet-3 & \cellcolor[HTML]{efefef}\textbf{15.5\%} & \cellcolor[HTML]{efefef}\textbf{23.8\%} & \cellcolor[HTML]{efefef}\textbf{40.9\%} & \cellcolor[HTML]{efefef}\textbf{98.6\%} \\ \bottomrule
            \end{tabular}%
        }
      \end{minipage}
  \end{minipage}
  \caption{\emph{Left.} Effective receptive field (ERF) comparison.
  FFNet has broader ERFs compared to Swin Transformer~\cite{liu2021swin} and ConvNeXt~\cite{liu2022convnet}, despite using the same kernel (window) size. \emph{Right.} Quantitative analysis on the ERF with the high-contribution area ratio $r$. A higher value of $r$ indicates a smoother distribution of high-contribution pixels, which implies a larger ERF.}
  \label{app_fig: erf}
\end{figure*}

As depicted in Fig.~\ref{app_fig: erf}, while high-contribution pixels in ConvNeXt and Swin are concentrated around the center of the input, FFNet’s high-contribution pixels are dispersed across a wider ERF.
These results demonstrate that our model more effectively utilizes convolutions with the same kernel size to consider a larger range of pixels for final predictions.
Moreover, FFNified attention exhibits a receptive field comparable to that of FastViT~\cite{vasu2023fastvit} and DeiT~\cite{touvron2021deit}, which employ global self-attention.
\emph{
These results indicate that despite using only convolution, FFNified attention harnesses large kernels effectively to capture global context.
These results are consistent with the examples in }Fig.~\ref{app_fig:FFNet_are_kvm}.

\section{Structural Re-parameterization} \label{app_sec: reparam}
\begin{figure}[t]
  \centering
  \includegraphics[width=\columnwidth]{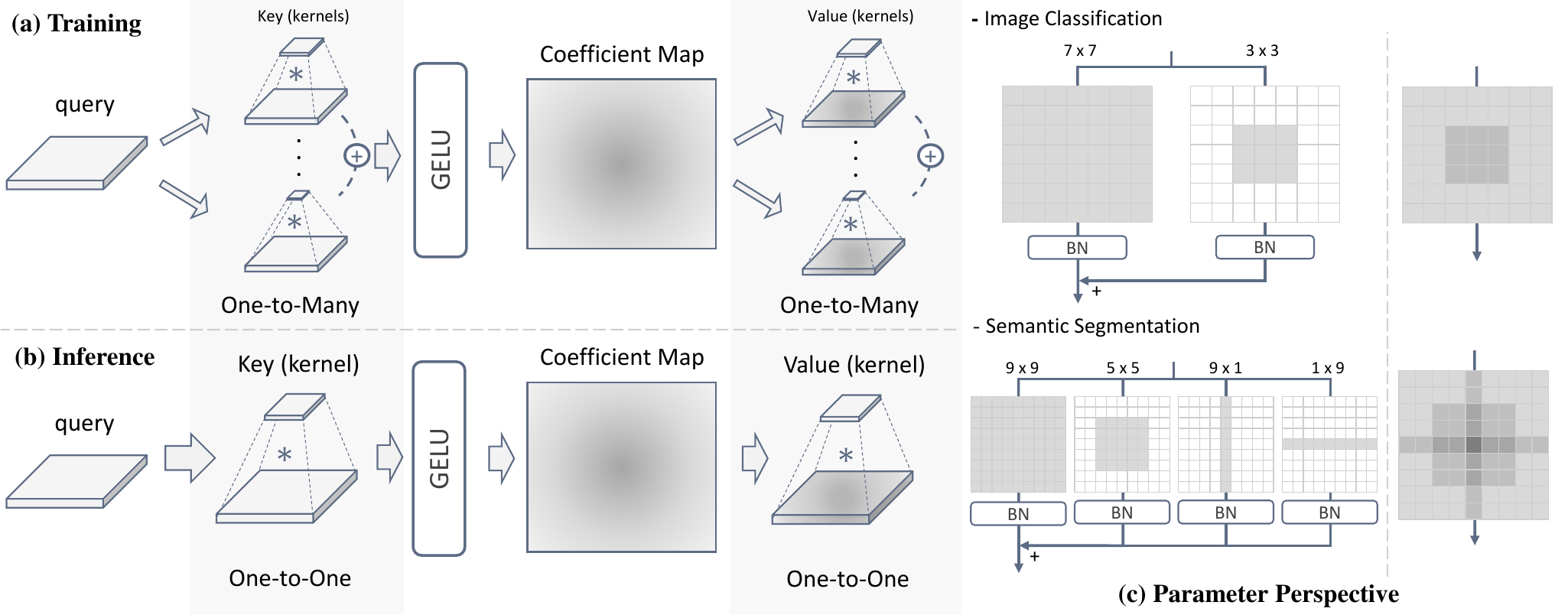}
  \caption{\small Visual explanation of re-parameterizing multi-branches for extracting multi-scale features.
  For a detailed discussion, please refer to Sec.~\ref{app_sec: reparam}.}
  \label{app_fig: reparam} 
\end{figure}

Structural re-parameterization~\cite{ding2021repvgg} is a methodology that transforms train-time multi-branch architectures into plain architectures through parameter transformation (Addition). 
We utilize this methodology to incorporate small kernels (e.g., 3$\times$3 or 5$\times$5) into larger ones (e.g., 7$\times$7 or 9$\times$9), as shown in Fig.~\ref{app_fig: reparam}. Additionally, for semantic segmentation model FFNet$_{seg}$, we add 9$\times$1 and 1$\times$9 convolution branches to effectively extract features of strip-like objects, such as human and tree. 
This approach enables our mixer to capture patterns of various scales and aspect ratios without adding inference latency. 

Interestingly, using multi-branches can be interpreted as one-to-many interactions, similar to the self-attention, leveraging multiple key-value pairs (kernels) for a single query (channel).
However, unlike self-attention, which typically considers a large number of key-values (e.g., 196 or 49 tokens) per query, our method uses only 2 or 4 key-values, making element-wise addition followed by GELU (“collaborative weighting”) more natural than softmax (“competitive weighting”).
Meanwhile, based on FFNified attention, leveraging softmax gating with multi-branches for context-dependent key-values~\cite{chen2020dynamic_conv} or handling multiple tasks~\cite{zhou2023rep_mode} presents an intriguing direction for future work.
As demonstrated in Tab.~\ref{app_tab:ablation_reparam}, employing the re-parameterization technique within our mixer design effectively boosts performance without additional inference latency.
However, this technique results in increased training time due to the added branches.

\begin{table}[t]
    \centering
    \caption{
        Comparison of FFNet models with and without multi-branches using re-parameterization technique when trained on ImageNet-1k. Train time refers to the wall clock time elapsed at the end of a training run, as measured using 8 RTX A6000 GPUs.
        }
        \label{app_tab:ablation_reparam}
        \tablestyle{3pt}{1.0}
        \begin{tabular}{l|c|c|c}
    \toprule
    Variant   & Ablation   & Train Time (hrs) & Top-1 
 Acc. (\%)  \\ 
    \midrule
    \multirow{2}{*}{FFNet-1} & plain architecture & 34.5 & 81.0 \\
                                & multi-branches w/ re-param. & 36.6 & 81.3 \\
    \midrule                                
    \multirow{2}{*}{FFNet-2} & plain architecture & 73.2 & 82.5 \\
                                & multi-branches w/ re-param. & 77.5 & 82.9 \\                                
    \bottomrule
    \end{tabular}%
\end{table}

\begin{table}[t]
    \centering
    \caption{
        Details about FFNet's architecture for ImageNet classification.
    The expansion ratio in the channel mixer of $\text{FFNet}_{\textit{seg}}$ is set to 4, while for all other variants (FFNet-1 $\sim$ 4), it is set to 3.
        }\label{app_tab:archi}
        \tablestyle{1.0pt}{0.9}
        \tiny
        \begin{tabular}{c c c c c c c}
    \toprule[1pt]
         Output size & Layer & FFNet-1 & FFNet-2 & FFNet-3 & FFNet-4 & $\text{FFNet}_{\textit{seg}}$ \\
         \midrule[0.5pt]
         $\frac{H}{4} \times \frac{W}{4}$ & stem & \makecell{$3\times3$, $64$, s = 2\\$3\times3$, $80$, s = 2}&\makecell{$3\times3$, $64$, s = 2\\$3\times3$, $88$, s = 2} & \makecell{$3\times3$, $64$, s = 2\\$3\times3$, $96$, s = 2} & \makecell{$3\times3$, $64$, s = 2\\$3\times3$, $128$, s = 2} & 
         \makecell{$3\times3$, $64$, s = 2\\$3\times3$, $64$, s = 1 \\$3\times3$, $96$, s = 1\\$3\times3$, $96$, s = 2 }\\
         \midrule[0.5pt]
         $\frac{H}{4} \times \frac{W}{4}$ & \makecell{token mixer \\ channel mixer} & $\begin{bmatrix}\setlength{\arraycolsep}{0.5pt}\begin{array}{c} 3\times3, 80 \\  3\times3, 80
         \end{array}\end{bmatrix}\times2$ & 
         $\begin{bmatrix}\setlength{\arraycolsep}{0.5pt}\begin{array}{c} 3\times3, 88 \\  7\times7, 88
         \end{array}\end{bmatrix}\times3$ &
         $\begin{bmatrix}\setlength{\arraycolsep}{0.5pt}\begin{array}{c} 3\times3, 96 \\  7\times7, 96
         \end{array}\end{bmatrix}\times4$ &
         $\begin{bmatrix}\setlength{\arraycolsep}{0.5pt}\begin{array}{c} 3\times3, 128 \\  7\times7, 128
         \end{array}\end{bmatrix}\times4$  & 
         $\begin{bmatrix}\setlength{\arraycolsep}{0.5pt}\begin{array}{c} 3\times3, 96 \\  3\times3, 96
         \end{array}\end{bmatrix}\times3$ \\
         \midrule[0.5pt]
         $\frac{H}{8} \times \frac{W}{8}$ & \makecell{token mixer \\ channel mixer} & 
         $\begin{bmatrix}\setlength{\arraycolsep}{0.5pt}\begin{array}{c} 3\times3, 160 \\  3\times3, 160
         \end{array}\end{bmatrix}\times2$ & 
         $\begin{bmatrix}\setlength{\arraycolsep}{0.5pt}\begin{array}{c} 3\times3, 176 \\  7\times7, 176
         \end{array}\end{bmatrix}\times3$ &
         $\begin{bmatrix}\setlength{\arraycolsep}{0.5pt}\begin{array}{c} 3\times3, 192 \\  7\times7, 192
         \end{array}\end{bmatrix}\times4$ &
         $\begin{bmatrix}\setlength{\arraycolsep}{0.5pt}\begin{array}{c} 3\times3, 256 \\  7\times7, 256
         \end{array}\end{bmatrix}\times4$ &
         $\begin{bmatrix}\setlength{\arraycolsep}{0.5pt}\begin{array}{c} 3\times3, 192 \\  3\times3, 192
         \end{array}\end{bmatrix}\times4$\\
         \midrule[0.5pt]
         $\frac{H}{16} \times \frac{W}{16}$ & \makecell{token mixer \\ channel mixer} & $\begin{bmatrix}\setlength{\arraycolsep}{0.5pt}\begin{array}{c} 7\times7, 320 \\ 3\times3, 320
         \end{array}\end{bmatrix}\times8$ & 
         $\begin{bmatrix}\setlength{\arraycolsep}{0.5pt}\begin{array}{c} 7\times7, 352 \\  7\times7, 352
         \end{array}\end{bmatrix}\times15$ &
         $\begin{bmatrix}\setlength{\arraycolsep}{0.5pt}\begin{array}{c} 7\times7, 384 \\  7\times7, 384
         \end{array}\end{bmatrix}\times22$ &
         $\begin{bmatrix}\setlength{\arraycolsep}{0.5pt}\begin{array}{c} 7\times7, 512 \\  7\times7, 512
         \end{array}\end{bmatrix}\times27$ &
         $\begin{bmatrix}\setlength{\arraycolsep}{0.5pt}\begin{array}{c} 9\times9, 384 \\  3\times3, 384
         \end{array}\end{bmatrix}\times18$\\
         \midrule[0.5pt]
         $\frac{H}{32} \times \frac{W}{32}$ & \makecell{token mixer \\ channel mixer} & $\begin{bmatrix}\setlength{\arraycolsep}{0.5pt}\begin{array}{c} 7\times7, 640 \\ 3\times3, 640
         \end{array}\end{bmatrix}\times2$ & 
         $\begin{bmatrix}\setlength{\arraycolsep}{0.5pt}\begin{array}{c} 7\times7, 704 \\  7\times7, 704
         \end{array}\end{bmatrix}\times3$ &
         $\begin{bmatrix}\setlength{\arraycolsep}{0.5pt}\begin{array}{c} 7\times7, 768 \\  7\times7, 768
         \end{array}\end{bmatrix}\times5$ &
         $\begin{bmatrix}\setlength{\arraycolsep}{0.5pt}\begin{array}{c} 7\times7, 1024 \\  7\times7, 1024
         \end{array}\end{bmatrix}\times3$ &
         $\begin{bmatrix}\setlength{\arraycolsep}{0.5pt}\begin{array}{c} 9\times9, 768 \\  3\times3, 768
         \end{array}\end{bmatrix}\times5$\\
         \midrule[0.5pt]
         $1\times1$ & head & \multicolumn{5}{c}{Fully-Connected Layer, 1000}\\
         \bottomrule[1pt]
    \end{tabular}
    
\end{table}

\section{Implementation details}
\label{app_sec: implementation details}
\subsection{Image Recognition} \label{app_sec: image recog.}
\noindent \textbf{Architecture.} The detailed architecture hyperparameters
are shown in Tab.~\ref{app_tab:archi}.
Given an input image, we first apply two 3$\times$3 strided convolutions to it.
Batch normalization and GELU are used after each 3$\times$3 convolution in stem.
Then, the tokens pass through four stages
of stacked FFNet blocks for hierarchical feature
extraction.
In the early stages, 3$\times$3 convolutions are primarily used to capture high-frequency local information, followed predominantly by 7$\times$7 convolutions in later stages for capturing low-frequency global information~\cite{si2022inceptionformer}.
For downsampling, we utilize a 7$\times$7 strided depthwise convolution followed by a 1$\times$1 convolution.
We also use LayerScale~\cite{touvron2021going} to make our deep models more stable during training.

For system-level segmentation comparison, we introduce $\text{FFNet}_{\textit{seg}}$, a model that exclusively utilizes FFNet blocks across both the backbone and head.
Recent works~\cite{Cai_2023efficientvit, guo2022segnext, zhao2017pyramid} demonstrate that long-range interactions and multi-scale features are crucial for competitive performance in segmentation models. Accordingly, to enlarge the receptive field, we employ 9$\times$9 kernels in the final two stages of the backbone and the head. Additionally, to effectively process features of objects with varying sizes, we combine small-kernel and strip convolutions~\cite{hou2020strip, peng2017largestrip} (9$\times$1, 1$\times$9) with large-kernel convolution, enhancing flexible feature extraction.
For this, when employing 7$\times$7 or 9$\times$9 convolutions, we leverage re-parameterization technique with the train-time multi-branches introduced in Appendix~\ref{app_sec: reparam}.
The kernel size for depthwise convolution in the channel mixer is set to 7$\times$7 for all models except FFNet-1 and $\text{FFNet}_{\textit{seg}}$, which use 3$\times$3 kernels.

\begin{figure}[t]
  \centering
  \includegraphics[width=\textwidth]{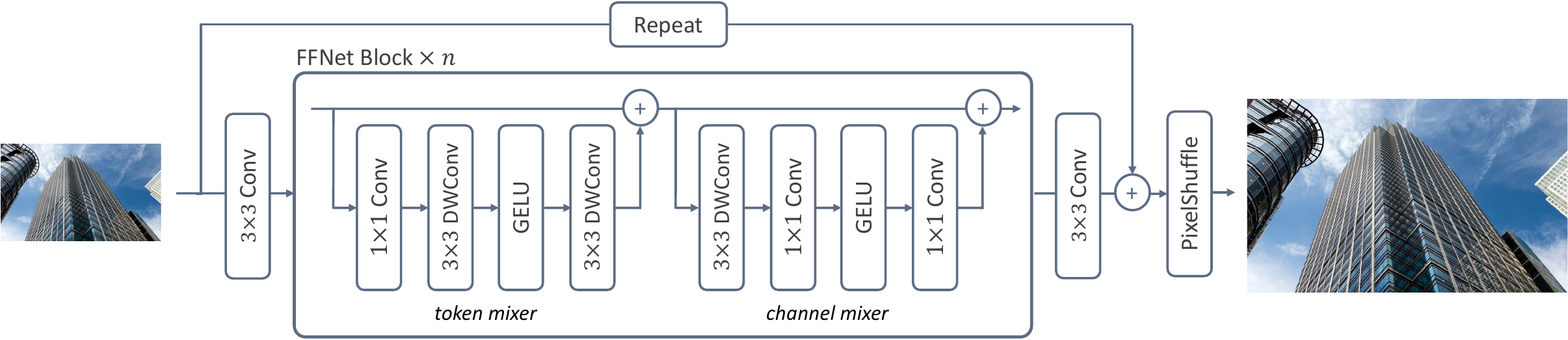}
  \caption{Overall architecture of our $\text{FFNet}{_\textit{sr}}$ for super-resolution task.}
  \label{app_fig: sr_architecture} 
\end{figure}

\subsection{Super-Resolution} \label{app_sec: sr-architecture}
\noindent \textbf{Architecture.} In this section, we explain the architecture of the proposed $\text{FFNet}{_\textit{sr}}$, adapted for the Super-Resolution~(SR) task, as shown in Fig.~\ref{app_fig: sr_architecture}.
$\text{FFNet}{_\textit{sr}}$ consists of three main components: the Shallow Feature Extractor~($H_{SF}$), the Deep Feature Extractor~($H_{DF}$), and the Upscaling Module~($H_{Rec}$).

The shallow feature extractor, $H_{SF}$, serves as the initial stage of the $\text{FFNet}_{\textit{sr}}$ architecture. 
It utilizes a 3$\times$3 convolution to process the Low-Resolution~(LR) input image capturing shallow features. 
The transformed shallow feature map has a channel size of 96, which is consistent across all variants.

The deep feature extractor, $H_{DF}$, refines the shallow features to produce finer deep features. 
Each $H_{DF}$ in $\text{FFNet}_{\textit{sr}}$-light and $\text{FFNet}_{\textit{sr}}$ consist of 36 and 48 FFNet Blocks respectively, followed by a final 3$\times$3 convolution. 
We set the kernel size of all depthwise convolutions to 3 to effectively extract high-frequency details and the expansion ratio of FFN to 2.
Importantly, we exclude Batch normalization from the FFNet blocks due to performance degradation, consistent with findings from EDSR~\cite{EDSR}.
The final 3$\times$3 convolution in $H_{DF}$ reduces the deep feature map's channels to match those of the long residual skip~\cite{RCAN}, allowing them to be added element-wise.

We utilize an anchor-based residual skip for the long residual skip, where the input LR image is repeated $r^{2}$ times along RGB channel-wise~\cite{ABPN, PCEVA}, where $r$ denotes the upscaling factor.
After passing through $H_{Rec}$, the repeated LR image becomes equivalent to a nearest-neighbor interpolated image, enabling the feature extractors to concentrate on recovering high-frequency details.

The upscaling module, $H_{Rec}$, takes the deep features and the long-residual skip, adds them together, and then upscales the result to reconstruct the High-Resolution (HR) image. 
It uses a single PixelShuffle~\cite{ESPCN} operation, which rearranges the channel dimensions into spatial dimensions in proportion to the $r$. 
This approach is more efficient and effective than the Transposed Convolution, which often results in checkerboard artifacts~\cite{ProSR}.

\noindent \textbf{Settings.} In each training batch, we randomly crop LR patches of size 64$\times$64 as input and augment the patches with random horizontal flips and rotations.
We use AdamW optimizer for training with a batch size of 64 for 1M iterations as follows previous study~\cite{SMFANet, lee2024implicit, lee2026rank}.
The initial learning rate is set to 0.001 and updated by the cosine annealing scheme.
All latency and memory usage are measured corresponding to an HR image of size 1280$\times$720.

\subsection{3D Semantic Segmentation} \label{app_sec: 3d segmentation}
\noindent \textbf{Architecture.} We adopt the serialization-based point cloud preprocessing introduced in PTv3~\cite{wu2023ptv3} and use a submanifold convolution with 5$\times$5$\times$5 kernel size for embeddings.
The encoder and decoder depths are set at [2, 2, 6, 2] and [1, 1, 1, 1], respectively, with channel numbers of [64, 128, 256, 512] for the encoder and [64, 64, 128, 256] for the decoder.
The expansion ratio for channel mixer is uniformly set at 4 across all blocks, with GELU as the activation function.
Furthermore, we incorporate batch normalization after each submanifold convolution, in addition to the pre-norm, to further stabilize the variance of the feature maps.

\noindent \textbf{Settings.} We train all models using a grid size of 0.02m, a batch size of 12, and a drop path rate of 0.3.
We employ the AdamW optimizer with an initial learning rate of $5 \times 10^{-3}$, adjusted via cosine decay, and a weight decay of 0.05.
Models are trained for 800 epochs on ScanNet series~\cite{dai2017scannet, rozenberszki2022language} and 3000 epochs on S3DIS~\cite{armeni2016s3dis} datasets. The data augmentation strategies employed are identical to those used in the PT series~\cite{wu2022point, wu2023ptv3}.

\begin{figure*}[t]
  \centering
  \includegraphics[width=0.9\textwidth]{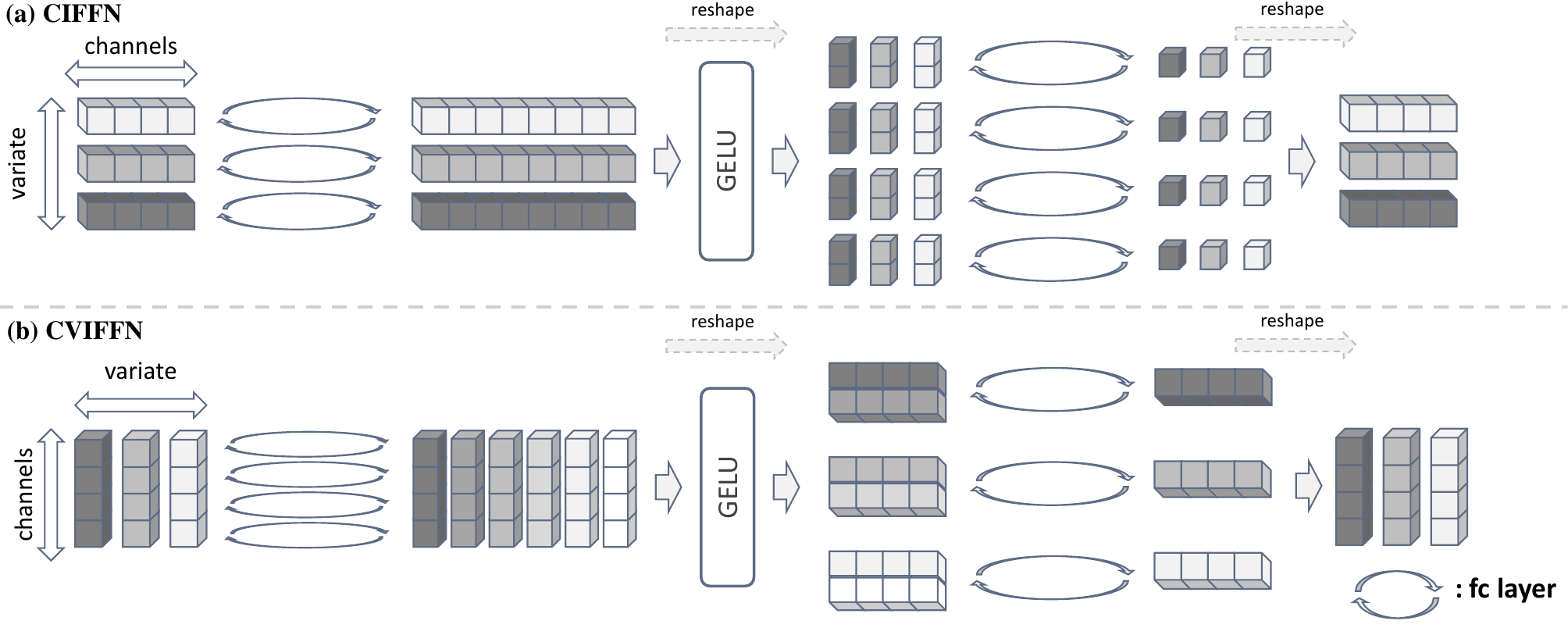}
  \includegraphics[width=0.65\textwidth]{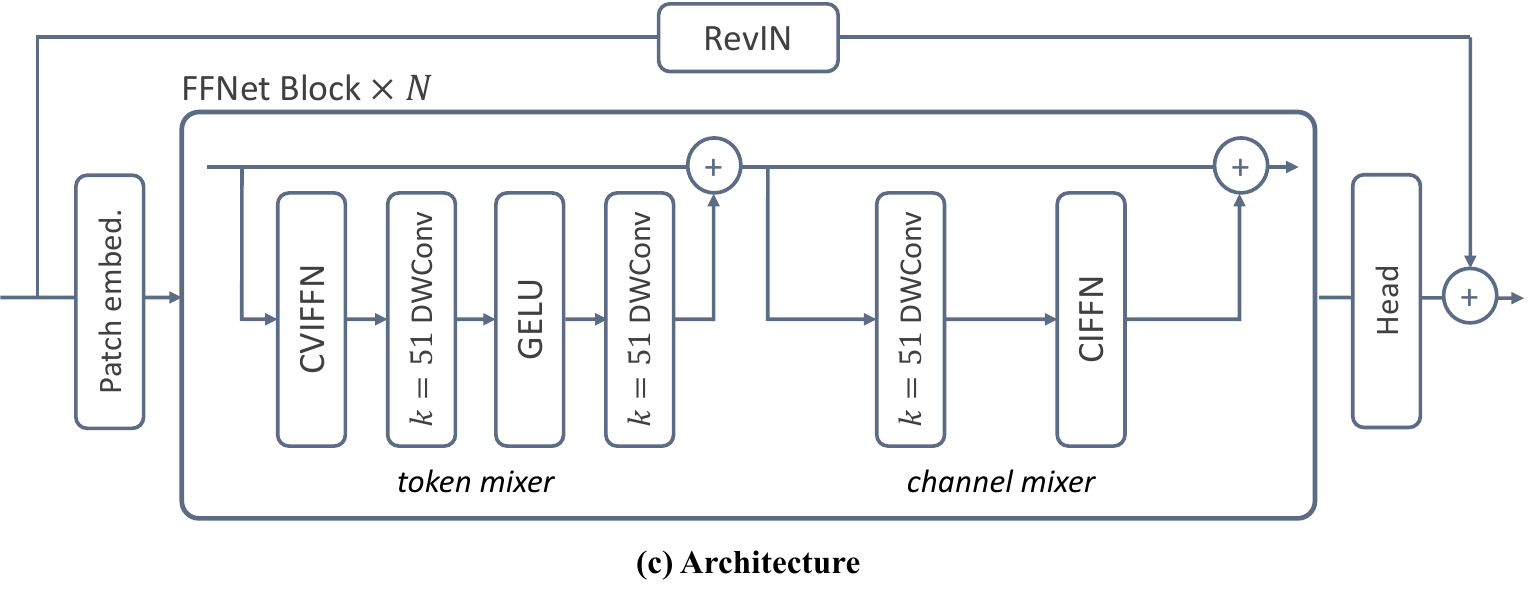}
  \caption{(a-b) Our proposed two types of FFN focus on channel interaction and cross-variable interaction via the first fully-connected layer, respectively. (c) After embedding, stacked FFNet blocks compute cross-time and cross-variable features. Then the final features are fed into the regression head.}
  \label{app_fig: ts_architecture} 
\end{figure*}

\begin{table}[t]
  \centering
  \caption{The overall information of the 8 long-term forecasting datasets.}\label{tab:ts-dataset}
  \begin{threeparttable}
  \renewcommand{\multirowsetup}{\centering}
  \setlength{\tabcolsep}{6.5pt}
  \tiny
  \begin{tabular}{l|c|c|c|c}
    \toprule
    Dataset &  Variable Number & Dataset Size & Sampling Frequency & Information \\
    \toprule
     ETTh (1, 2)~\cite{zhou2021informer} & 7  & (8545, 2881, 2881) & 1 hour & Electricity\\
     \midrule
     ETTm (1, 2)~\cite{zhou2021informer} & 7  & (34465, 11521, 11521) & 15 mins & Electricity\\
    \midrule
    Exchange~\cite{Autoformer} & 8  & (5120, 665, 1422) & 1 day & Economy \\
    \midrule
    Weather~\cite{Autoformer} & 21  & (36792, 5271, 10540) & 10 mins & Weather\\
    \midrule
    ECL~\cite{Autoformer} & 321  & (18317, 2633, 5261) & 1 hour & Electricity \\
    \midrule
    Traffic~\cite{Autoformer} & 862  & (12185, 1757, 3509) & 1 hour & Transportation \\
    \bottomrule
    \end{tabular}
  \end{threeparttable}
\end{table}

\subsection{Time Series Forecasting} \label{app_sec: forecasting}
\noindent \textbf{Architecture.} We introduce two FFN variants tailored for time series data, which maintain the structural integrity of our FFNet block design while effectively computing cross-time and cross-variable interactions.
Given the limitations of vanilla FFNs that operate solely along the channel axis, in a multivariate setting with both temporal and variable axes, we propose CVIFFN (Cross-Variable Interaction focused FFN) and CIFFN (Channel Interaction focused FFN) to capture cross-variable dependencies and channel information, respectively.
Details of these modules, including PyTorch-like code and intuitive visual explanations, can be found in Algorithm \ref{alg:time series CVIFFN}, \ref{alg:time series CIFFN}, and Fig. \ref{app_fig: ts_architecture}.
Both CVIFFN and CIFFN capture cross-variable and channel information through the first Fully-Connected (FC) layer respectively, then aggregate features across both axes using appropriate group$\&$reshape operations and  a second FC layer.
Thus, our FFNet block for time series replaces the basic block's FFN with CIFFN and the token mixer’s 1$\times$1 convolution with CVIFFN (see Fig.~\ref{app_fig: ts_architecture} (c)).

\begin{algorithm}[t]
\caption{PyTorch-style Code for Cross-Variable Interaction focused FFN}
\label{alg:time series CVIFFN}
\definecolor{codeblue}{rgb}{0.25,0.5,0.5}
\definecolor{codekw}{rgb}{0.85, 0.18, 0.50}
\lstset{
  backgroundcolor=\color{white},
  basicstyle=\fontsize{7.5pt}{7.5pt}\ttfamily\selectfont,
  columns=fullflexible,
  breaklines=true,
  captionpos=b,
  commentstyle=\fontsize{7.5pt}{7.5pt}\color{codeblue},
  keywordstyle=\fontsize{7.5pt}{7.5pt}\color{codekw},
}
\begin{lstlisting}[language=python]
import torch.nn as nn

class CVIFFN(nn.Module):
    """ Cross-Variable Interaction focused Feed-Forward Network """
    def __init__(self, d_model, e_r, n_vars, drop1, drop2):
        super().__init__()
        d_hidden = d_model * e_r
        self.e_r = e_r 

        self.fc1 = nn.Conv1d(in_channels=n_vars * d_model, out_channels=n_vars * d_hidden, kernel_size=1, stride=1, padding=0, groups=d_model)
        self.act = nn.GELU()
        self.fc2 = nn.Conv1d(in_channels=n_vars * d_hidden, out_channels=n_vars * d_model, kernel_size=1, stride=1, padding=0, groups=n_vars)
        
        self.drop1 = nn.Dropout(drop1)
        self.drop2 = nn.Dropout(drop2) 

    def forward(self,x):
        B, M, D, N = x.shape
        
        x = x.permute(0, 2, 1, 3)
        x = x.reshape(B, D * M, N)
        x = self.drop1(self.fc1(x))
        x = self.act(x)   
        x = x.reshape(B, D, self.e_r * M, N)
        x = x.permute(0, 2, 1, 3)   
        x = x.reshape(B, M * self.e_r * D, N)
        x = self.drop2(self.fc2(x))  
        x = x.reshape(B, M, D, N)

        return x
\end{lstlisting}
\end{algorithm}

\begin{algorithm}[t]
\caption{PyTorch-style Code for Channel Interaction focused FFN}
\label{alg:time series CIFFN}
\definecolor{codeblue}{rgb}{0.25,0.5,0.5}
\definecolor{codekw}{rgb}{0.85, 0.18, 0.50}
\lstset{
  backgroundcolor=\color{white},
  basicstyle=\fontsize{7.5pt}{7.5pt}\ttfamily\selectfont,
  columns=fullflexible,
  breaklines=true,
  captionpos=b,
  commentstyle=\fontsize{7.5pt}{7.5pt}\color{codeblue},
  keywordstyle=\fontsize{7.5pt}{7.5pt}\color{codekw},
}
\begin{lstlisting}[language=python]
import torch.nn as nn

class CIFFN(nn.Module):
    """ Channel Interaction focused Feed-Forward Network """
    def __init__(self, d_model, e_r, n_vars, drop1, drop2):
        super().__init__()
        self.d_hidden = d_model * e_r 

        self.fc1 = nn.Conv1d(in_channels=n_vars * d_model, out_channels=n_vars * self.d_hidden, kernel_size=1, stride=1, padding=0, groups=n_vars)
        self.act = nn.GELU()
        self.fc2 = nn.Conv1d(in_channels=n_vars * self.d_hidden, out_channels=n_vars * d_model, kernel_size=1, stride=1, padding=0, groups=d_model)
        
        self.drop1 = nn.Dropout(drop1)
        self.drop2 = nn.Dropout(drop2) 

    def forward(self,x):
        B, M, D, N = x.shape
        
        x = x.reshape(B, M * D, N)
        x = self.drop1(self.fc1(x))
        x = self.act(x)   
        x = x.reshape(B, M, self.d_hidden, N)
        x = x.permute(0, 2, 1, 3)   
        x = x.reshape(B, self.d_hidden * M, N)
        x = self.drop2(self.fc2(x))  
        x = x.reshape(B, D, M, N)
        x = x.permute(0, 2, 1, 3)

        return x
\end{lstlisting}
\end{algorithm}

By default, our model includes 1 FFNet block with a hidden dimension of 64 and an expansion ratio of 12. For patch embedding, the patch size and stride are set at 4 and 2, respectively. 
For the Exchange, ETTh1, and ETTh2 datasets, we reduce the expansion ratio to 1, 2, and 8, respectively, to mitigate overfitting.
To address distribution shift problem, we also utilize RevIN~\cite{kim2022reversible}.
Finally, for features $O \in \mathbb{R}^{M \times D \times N}$ , we flatten them along the channel $D$ and token $N$ axes, and then compute the final prediction using a linear layer with output channels equal to the prediction length.

\noindent \textbf{Settings.} We train our model with the L2 loss, using Adam optimizer with an initial learning rate of $10^{-4}$.
The standard training procedure is 100 epochs, incorporating proper early stopping.
Following~\cite{liu2024itransformer}, we fix lookback length 96 for all the datasets and
all experiments are repeated three times.
We also borrow from the baseline results from iTransformer~\cite{liu2024itransformer}.
Detailed dataset descriptions are summarized in Tab.~\ref{tab:ts-dataset}.

\section{Qualitative Results}
\subsection{Super-Resolution}
\label{app_sec: sr_vis}
\begin{figure*}[t]
  \centering
  \includegraphics[width=0.8\columnwidth]{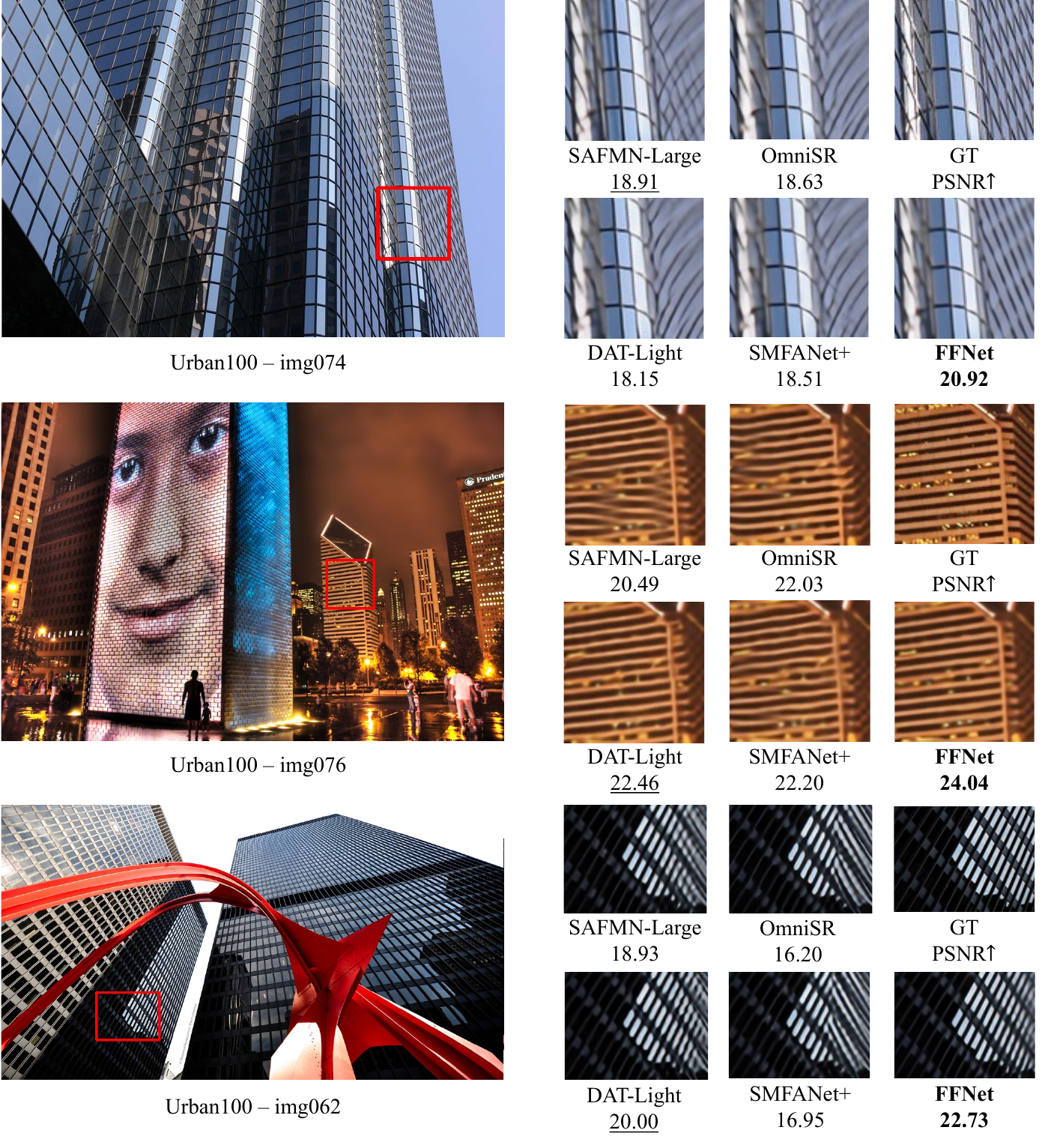}
  \caption{Visual comparison for $\times$4 SR models. The patches for comparison are marked with red boxes. Zoom-in for better visibility.
  }
  \label{app_fig:sr_visual_results} 
\end{figure*}

In this section, we present visual comparisons of our $\text{FFNet}_{\textit{sr}}$ model against state-of-the-art SR models at $\times$4 scale.
Figure \ref{app_fig:sr_visual_results} zooms in on each SR result to emphasize the differences in image sharpness and fine-grained detail preservation, providing corresponding PSNR values. 
These visualizations underscore the exceptional performance of $\text{FFNet}_{\textit{sr}}$ in the SR task, demonstrating task-generality of the MetaMixer-based convolutional framework.

\subsection{ImageNet Generation}
\begin{figure*}[t]
  \centering
  \includegraphics[width=0.9\columnwidth]{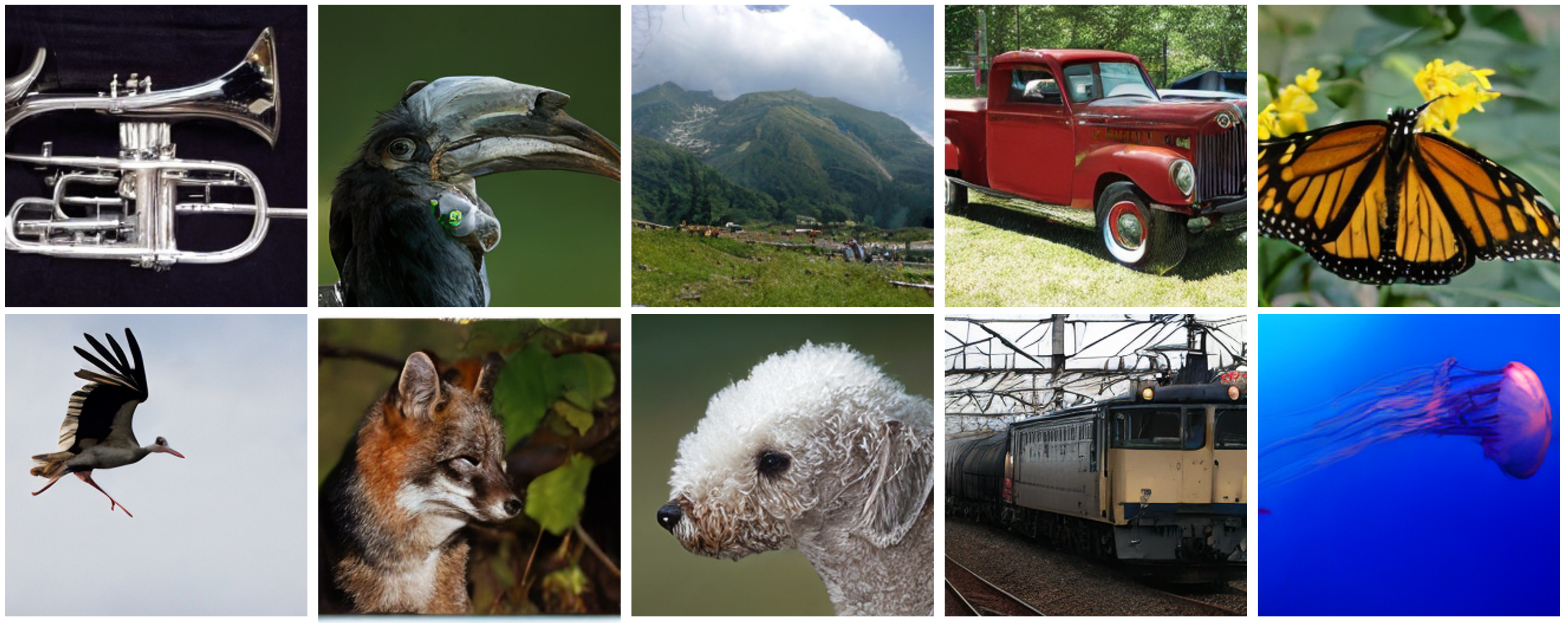}
  \caption{Selected samples generated using our U-FFNet ImageNet-256 model with classifier-free guidance~\cite{ho2022classifier}.
  }
  \label{app_fig:imagenet_samples}
\end{figure*}

We provide the generated samples of latent diffusion model with our hybrid mixer design in Fig.~\ref{app_fig:imagenet_samples}.
Our results confirm that our proposed mixer design is also effective in generative modeling.

\section{Limitations} \label{app_sec: limitations}
While our proposed MetaMixer-based convolutional mixer demonstrates remarkable performance across multiple tasks, showcasing that the MetaMixer itself is crucial rather than specific modules like attention, its effectiveness in more diverse scenarios, such as large-scale datasets like ImageNet-21k, remains unproven.
Moreover, we acknowledge that the performance of our model is partially improved with the help of previous techniques.
However, we hope that our unified perspectives and possibilities we present will serve as a starting point for future mixer design space and strengthen the systematic approach to model development.

\clearpage

%
%
\bibliographystyle{splncs04}
\bibliography{main}
\end{document}